\documentclass{article}

\newcommand{\hlMethod}[1]{\textit{\textbf{#1}}:}

\usepackage{microtype}
\usepackage{graphicx}
\usepackage{booktabs} 
\usepackage{amsfonts}

\usepackage{hyperref}

\usepackage{stfloats}
\usepackage{mathtools}
\usepackage[accepted]{icml2021}
\usepackage{paralist}
\usepackage{tikz}
\usepackage{amsmath}
\usepackage{xcolor}
\usetikzlibrary{positioning}
\usepackage{adjustbox}
\usetikzlibrary{shapes,arrows,calc}
\usepackage{algorithm}
\usepackage{algorithmic}
\usepackage{float}
\usepackage{graphicx}
\usepackage{subcaption}
\usepackage{sidecap}
\usepackage{tabularx}

\usepackage[skip=-2pt]{caption} 

\icmltitlerunning{Dopamine-driven synaptic credit assignment in neural networks}

\begin{document}

\twocolumn[
\icmltitle{Dopamine-driven synaptic credit assignment in neural networks}



\icmlsetsymbol{equal}{*}

\begin{icmlauthorlist}
\icmlauthor{Saranraj Nambusubramaniyan}{ir}
\icmlauthor{Shervin Safavi}{max}
\icmlauthor{Raja Guru}{equal,ir}
\icmlauthor{Andreas Knoblauch}{equal,as}

\end{icmlauthorlist}

\icmlaffiliation{ir}{Indian center for Robotics Innovation and Smart-intelligence, Chennai, India}
\icmlaffiliation{max}{Max Planck Institute for Biological Cybernetics, Tuebingen, Germany}
\icmlaffiliation{as}{Informatics Faculty, Albstadt-Sigmaringen University, Germany}

\icmlcorrespondingauthor{Saranraj Nambusubramaniyan}{saran@iris-i.in}

\icmlkeywords{Recurrent Neural Networks, Neuro-inspired computing, Weight perturbation learning, Credit assignment, Weight symmetry, Update locking, Chaotic time-series prediction, Derivative-free learning}

\vskip 0.3in
]




\begin{figure*}[b!]
\footnotesize
\rule{\textwidth}{0.4pt}
$^{1}$Indian center for Robotics Innovation and Smart-intelligence, Chennai, India.
$^{2}$Max Planck Institute for Biological Cybernetics, Tuebingen, Germany.
$^{3}$Informatics Faculty, Albstadt-Sigmaringen University, Germany.
Correspondence: saran@iris-i.in
\end{figure*}

\begin{abstract}
Solving the synaptic Credit Assignment Problem (CAP) is central to learning in both biological and artificial neural systems. Finding an optimal solution for synaptic CAP means setting the synaptic weights that assign credit to each neuron for influencing the final output and behavior of neural networks or animals. Gradient-based methods solve this problem in artificial neural networks using back-propagation, however, not in the most efficient way. For instance, back-propagation requires a chain of top-down gradient computations. This leads to an expensive optimization process in terms of computing power and memory linked with well-known weight transport and update locking problems. To address these shortcomings, we take a NeuroAI approach and draw inspiration from neural Reinforcement Learning to develop a derivative-free optimizer for training neural networks, \textit{Dopamine}. Dopamine is developed for Weight Perturbation (WP) learning that exploits stochastic updating of weights towards optima. It achieves this by minimizing the regret, a form of Reward Prediction Error (RPE) between the expected outcome from the perturbed model and the actual outcome from the unperturbed model. We use this RPE to adjust the learning rate in the network (i.e., creating an adaptive learning rate strategy, similar to the role of dopamine in the brain). We tested the Dopamine optimizer for training multi-layered perceptrons for XOR tasks, and recurrent neural networks for chaotic time series forecasting. Dopamine-trained models demonstrate accelerated convergence and outperform standard WP, and give comparable performance to gradient-based algorithms, while consuming significantly less computation and memory. Overall, the Dopamine optimizer not only finds robust solutions and comparable performance to the state-of-the-art Machine Learning optimizers but is also neurobiologically more plausible.

\end{abstract}

\section{Introduction}
\label{submission}
Solving the synaptic Credit Assignment Problem (CAP) is central to learning in both biological and artificial neural networks \citep{williams1989learning,lillicrap2020backpropagation,crick1989recent}. 
The solution for synaptic CAP is determined by setting the synaptic weights that assign credit to each neuron for influencing the final output and behavior of animals or artificial networks. The optimal assignment of these weights leads to the best performance of the network.
In artificial networks, CAP is often solved by gradient-based Back-Propagation (BP) methods \citep{schmidhuber2015deep} (e.g., different variants of stochastic gradient descent), most notably, Adaptive Moment Estimation or Adam \citep{kingma2014adam}.
In BP, the gradient of a loss function concerning the network parameters for a given input-output pair is computed through a chain of top-down computations. 
However, this may not be an efficient way to solve CAP, as it can be expensive in terms of computing power to estimate the gradient along all backward steps and memory to keep the gradient until the parameter is updated (i.e. weight transport and update locking problems) \citep{williams1995gradient,liu2021cell}.

The problems with gradient-based methods are exacerbated when training Recurrent Neural Networks (RNNs) and their variants (e.g., GRU and LSTM) that require solving the long-term CAP that may lead to vanishing or exploding gradients~\citep{elman1990finding}.
RNNs allowed developing models capable of processing long temporal dependencies of sequential data over variable timescales such as in language recognition/synthesis \citep{bowman2015generating}, time-series prediction \citep{chung2014empirical, torres2021deep}, and music composition \citep{boulanger2012modeling} and have extensively been used as proxies for biological neural networks to study a wide range of cognitive functions including learning, memory, and adaptation \cite{glaser2019roles, kawato1988adaptation, zipser1991recurrent, kroes2012dynamic,bianchi2020bio, quax2020adaptive,landi2021working,zipser1991recurrent, sussillo2014neural, kao2019considerations, liu2023brain}.
The conventional optimizers for training RNNs are gradient-based algorithms such as Back-Propagation Through Time (BPTT) and Real-Time Recurrent Learning (RTRL) for batch and online learning respectively. 
However, both methods come with caveats: BPTT requires inputs and hidden states to be saved in memory, making it impractical for longer sequences \citep{hochreiter2001gradient}, whereas RTRL suffers from high computational cost per update \citep{subramoney2023efficient}. 
Some of these problems may be mitigated by Truncated BPTT \cite{williams1990efficient,jaeger2002tutorial}, but it does not fully resolve the problems related to efficiently solving CAP. 

In addition, BP is considered to be biologically implausible \citep{lillicrap2020backpropagation} for instance due to (i) non-local update rules (synaptic CAP), (ii) weight symmetry \citep{lillicrap2014random, liao2016important}, i.e., forcing symmetric constraint on feedforward and feedback weights, and (iii) forward update and backward locks \cite{jaderberg2017decoupled}, since BP requires buffering the inputs and freezing the connection strengths. Such update rules do not elucidate the biological basis of local as well as online learning \cite{lillicrap2020backpropagation,stiber2005spike,bengio2015towards}.

This motivated the search for more efficient alternative ways to solve synaptic CAP, inspired by learning mechanisms in the brain.
Some of these methods include \textit{Target Propagation} \citep{ackley1985learning, le1986learning} addressing the local synaptic CAP, \textit{Feedback Alignment} \citep[FA,][]{lillicrap2016random, nokland2016direct, flugel2023feed}
designed for weight symmetry problem and \textit{Decoupled Neural Interface} \citep[DNI,][]{jaderberg2017decoupled} specifically tailored for locking problems in deep networks.
At the same time, methods inspired by brain's synaptic plasticity rules \citep{lazar2009sorn,nambusubramaniyan2021sorn} and neuromodulatory signals (a key mechanism for solving CAP in brain \citep{dayan2012twenty}), combined with Reinforcement Learning (RL)\citep{liu2021cell,hamid2021wave,gupta2021structural}, gave rise to reward-based learning algorithms, generally called three-factor learning~\citep{fremaux2016neuromodulated,gerstner2018eligibility}. 
Three-factor learning combines a local Hebbian learning rule (i.e., computing local gradients) with a global error signal inspired by neuromodulatory mechanisms providing an approximation of gradient descent. However, these algorithms often learn slower than BPTT failing to address all the above mentioned caveats of gradients based learning including computational efficiency.

\begin{figure*}[ht]
\centering
\tikzstyle{block} = [rectangle, draw, text width=4em, text centered, rounded corners, minimum height=1em, minimum width=0.5em]
\tikzstyle{block1} = [circle, draw, 
text width=1em, text centered, minimum height=1em]
\tikzstyle{line} = [draw, -latex']
\tikzstyle{xi} = [ draw, text width=1em, text centered, aspect=1, inner sep=0pt, minimum height=1em]

\begin{minipage}[t]{0.48\textwidth} 
\centering
\begin{tikzpicture}[node distance = 2cm]
        \node (x) at (0,0) {};
	\node (r) at (0,0) {};
	\node (xi) at (0,0) {};
    \node [block] (gl) {$g^l$};
	\node [block, right of=gl, node distance=2.5cm] (gl1) {$g^{l+1}$ };
	\node [block, left of=gl1, node distance=2.5cm] (gl) {$g^l$};
	\node [left of=gl, node distance=1.5cm] (x) {$x_t$};
	\node [right of=gl1, node distance=1.6cm] (r) {\color{red} $\mathcal{R}_t$};
	
	\path [line] (x) -- (gl);
	\draw [dotted, -latex'] ($(gl1.east)!0.5!(gl1.south east)$) -- ++(0.5,0);
	\draw [line, -latex'] ($(gl1.east)!0.5!(gl1.east)$) -- ++(0.5,0);
	\draw [line, dotted, latex-] ($(gl1.west)!0.5!(gl1.south west)$) -- ++(-0.5,0) |- ($(gl.east)!0.5!(gl.south east)$);
	\path [line] (gl) -- (gl1);
	
	\node [ below = 1em and 0cm of gl1] (xi) {$\xi^{l+1}$};
	
	\draw [line, dotted] (xi.north) to[out=90,in=270] ($(gl1.south)!0.5!(gl1.south west)$);
	\draw [line, dotted, red] (xi.north) to[out=90,in=270] ($(gl1.south)!0.5!(gl1.south east)$);
	\node [ below = 1em and 0cm of gl] (xi) {$\xi^l$};
	
	\draw [line, dotted] (xi.north) to[out=90,in=270] ($(gl.south)!0.5!(gl.south west)$);
	\draw [line, dotted, red] (xi.north) to[out=90,in=270] ($(gl.south)!0.5!(gl.south east)$);
\end{tikzpicture}

\vspace{0.3cm} 
\caption{{Multi-layered feed-forward module $g$.} $g$ denotes nonlinearity and $x$ denotes the input to the network. $y(t)=g(x_t);l \in [1,L]$; $L=2$. Solid arrows indicate forward data flow with unperturbed parameters, while dotted arrows represent flow with perturbed parameters. Red dotted arrows illustrate the parallel, layerwise parameter update step.}
\vspace{-0.5cm}
\label{fig:ffnetwork}
\end{minipage}
\hfill
\begin{minipage}[t]{0.48\textwidth}
\centering
\begin{tikzpicture}[node distance = 1.5cm]
        \node (x) at (0,0) {};
	\node (x1) at (0,0) {};
	\node (x2) at (0,0) {};
	\node (r) at (0,0) {};
	\node (xi) at (0,0) {};
	
	\node [block1] (gl2) at (0,0) {$g_{i}$};
	\node [block1, left of=gl2] (gl1) {$g_{i}$};
	\node [block1, left of=gl1, node distance=1.5cm] (gl) {$g_{i}$};
	\node [above of=gl, node distance=1cm] (x) {$x_t,h_{t-1}$};
	\node [above of=gl1, node distance=1cm] (x1) {$x_{t+1},h_{t}$};
	\node [right of=gl2, node distance=1.3cm] (r) {\color{red} $\mathcal{R}_{\tau}$};
	\node [above of=gl2, node distance=1cm] (x2) {$x_{t+2},h_{t+1}$};
	
	\path [line] (x) -- (gl);
	\path [line] (x1) -- (gl1);
	\path [line] (gl1) -- (gl2);
	\path [line] (x2) -- (gl2);
	\draw [dotted, -latex'] ($(gl2.east)!0.5!(gl2.south east)$) -- ++(0.5,0);
	\draw [line, -latex'] ($(gl2.east)!0.5!(gl2.east)$) -- ++(0.5,0);
	\draw [line, dotted, latex-] ($(gl2.west)!0.5!(gl2.south west)$) -- ++(-0.5,0) |- ($(gl1.east)!0.5!(gl1.south east)$);
	\path [line] (gl) -- (gl1);
	\draw [line, dotted, latex-] ($(gl1.west)!0.5!(gl1.south west)$) -- ++(-0.5,0) |- ($(gl.east)!0.5!(gl.south east)$);
	
	\node [left of=gl, node distance=1.3cm] (xi) {$\xi^l$};
	\draw [line, dotted] (xi.east) to[out=90,in=180] ($(gl.west)!0.5!(gl.south west)$);
	\draw [line, dotted, red] (xi.east) to[out=90,in=180] ($(gl.west)!0.5!(gl.north west)$);
	
	\node [below of=gl, node distance=1cm] (r1) {$\mathcal{R}_t$};
	\path [line] (gl) -- (r1);
	\node [below of=gl1, node distance=1cm] (r2) {$\mathcal{R}_{t+1}$};
	\path [line] (gl1) -- (r2);
	\node [below of=gl2, node distance=1cm] (r3) {$\mathcal{R}_{t+2}$};
	\path [line] (gl2) -- (r3);
\end{tikzpicture}

\vspace{0.3cm} 
\caption{Recurrent core module $g_i$ unrolled over time.  $h$ represents the recurrent state. $y_t,h_t=g(x_t,h_{t-1});$ $g_i=g \ \ \forall_i;\tau=[t,T],T=3$. Solid arrows denote the temporal data flow through neuron $g_i$ and the red dotted arrows illustrate the time-independent parameter update step. }
\vspace{-0.5cm}
\label{fig:rnnnetwork}
\end{minipage}
\end{figure*}

To remedy these problems, here we draw inspiration from
neural Reinforcement Learning
\citep[][]{urbanczik2014learning,coddington2023mesolimbic}
to develop a neurobiologically plausible, and derivative-free optimizer for training neural networks. 
Specifically, we introduce the new optimizer, \emph{Dopamine}, which is inspired by Weight Perturbation (WP) learning \cite{jabri1991weight, dembo1990model,zuge2023weight}.
\emph{Dopamine} is computationally more efficient than gradient-based methods and provides comparable performance to the state-of-the-art optimizers in machine learning.
In \emph{Dopamine}, we exploit stochastic updating of the weights \citep{urbanczik2014learning} to build a form of Reward prediction error signal,
and the dynamic change of this signal (a moving average) to make it an adaptive process.
This optimizer is grounded in a few modest assumptions on biologically observed phenomena such as dopamine-driven hedonistic learning \citep{montague1996framework,schultz2002getting, menzel2001searching, seung2003learning}, spatio-temporally asynchronous dopamine signaling for credit assignment \citep{hamid2019dopamine}, stochasticity in synaptic efficacy \citep{korn1986probabilistic, mason1991synaptic, manwani2001detecting} and modulation of learning rate by reward signals \citep{urbanczik2014learning,coddington2023mesolimbic}. Hence, the proposed algorithm mitigates the aforementioned issues by
\begin{inparaenum}[(i)]
\item leveraging a global learning signal to acquire ``local error information" while adhering to locality constraints (CAP),
\item ensuring that feedforward and feedback weights are not identical by perturbing the feedforward weights to get reward signal (weight transport),
\item being gradient-free; it updates the parameters asynchronously during the backward pass, eliminating the need to freeze weights (locking problems), and additionally,
\item by exploiting the reward gradient, it enforces Reinforcement Learning through an adaptive learning rate that enables accelerated convergence.
\end{inparaenum}

\begin{figure}
    \begin{minipage}{0.6\textwidth} 
        \begin{subfigure}[b]{0.4\textwidth}
            \centering
            \caption*{}
            \includegraphics[width=\textwidth, height=\textwidth]{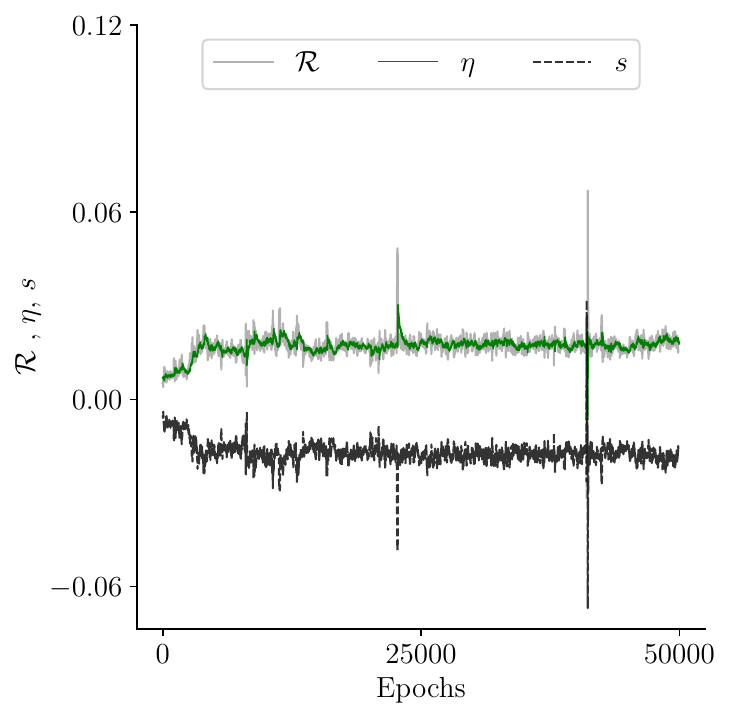}
        \end{subfigure}
        \hspace{-0.5em}
        \begin{subfigure}[b]{0.4\textwidth}
            \caption*{}
            \includegraphics[width=\textwidth, height=\textwidth]{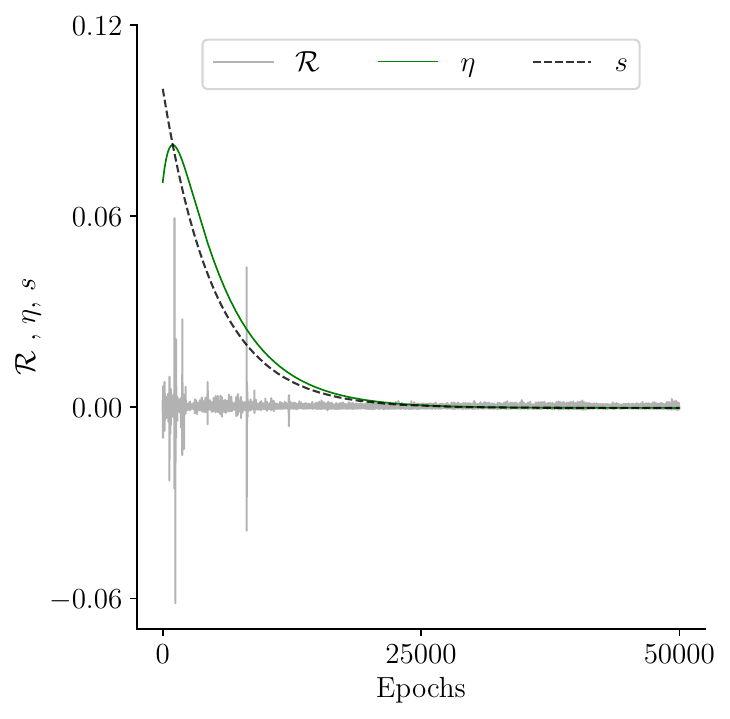}
        \end{subfigure}
    \end{minipage}%
    \hspace{-5em}
    \begin{minipage}{0.48\textwidth} 
    \centering
    \vspace{0.3cm}
        \caption{RPE $\mathcal{R}$ modulates learning rate $\eta$ while solving XOR-classification problem. \textit{Left}: \textit{Dopamine-1}: Learning rate follows the RPE($\mathcal{R}$) as observed in biological reinforcement learning \citep{urbanczik2014learning,coddington2023mesolimbic}. \textit{Right}: \textit{Dopamine-2}: Exponentially decays the learning rate towards $\mathcal{R}$, thereby avoiding abrupt parameter updates that might arise during the initial training stages. See \autoref{fig:betas_etas} for more details on the influence of moving average coefficients $\beta_s$ and $\beta_\eta$ and also the initial value of the auxiliary variable $s_0$ and the learning rate $\eta_0$. See hyperparameters details at \autoref{tab:xor_hyperparam_ablation} for Dopamine 1c (\textit{Left}) and Dopamine 2b (\textit{Right}).
        \vspace{-16pt}
        }        
        \label{fig:dopamine_update_rules}
    \end{minipage}
\end{figure}
\section{Related work}

Here we briefly review the alternative methods for BP, their advantages, and potential limitations.

\hlMethod{Target Propagation (TP)}
TP is a layerwise local learning approach motivated by biological learning \citep{le1986learning,lillicrap2020backpropagation}. TP and its variants \cite{bengio2014auto, lee2015difference, bartunov2018assessing, podlaski2020biological, shibuya2023fixed} propagate target activations instead of errors and update the weights of each layer to get closer to the target activation.
Nevertheless, effectively capturing the temporal dependencies in sequential data with TP is challenging \cite{bengio1994learning}. TP Through Time \cite{manchev2020target} addresses the issue of capturing the temporal dependencies in sequential data by approximating the inverses for gradient descent. However, it remains unclear whether the reverse layers truly learn the layer inverses during the training process. Moreover, this method involves additional computational complexity and the associated costs of learning approximate inverses \citep{roulet2021target}.

\hlMethod{Feedback Alignment (FA)} FA \citep{lillicrap2016random}, also known as random back-propagation \citep{baldi2018learning}, resolves the weight transport problem by using a fixed random connectivity matrix for error propagation. Direct FA \citep{nokland2016direct, crafton2019direct,refinetti2021align} on the other hand uses the product of the output error with the random feedback matrix for each hidden layer of the network. Furthermore, the Direct Random Target Projection (DRTP) method suggests that the error sign alone is sufficient to obtain the information about the modulatory signal and, thereby, does not require a backward pass \cite{flugel2023feed}. FA and its variants are argued to be biologically plausible since the error signal is almost local, and different weights are used for forward and backward passes. 

\hlMethod{Decoupled Neural Interface (DNI)}
DNI \citep{jaderberg2017decoupled} uses an error locality approach to remedy significant memory and computational overhead. 
To address the update-locking problem, DNI detaches the layers in the network and employs layerwise objective functions, i.e., the forward pass involves approximating the global loss gradient using the layer activations, and the parameters are updated instantaneously, rejecting the need for a backward pass~\citep{flugel2023feed,czarnecki2017understanding}. In RNNs, the synthetic gradients are evaluated by unrolling the hidden states over time.
Furthermore, DNI can converge to the optimal solution for non-convex problems.
Followed by a delayed error feedforward method, an error-driven input modulation method \citep{dellaferrera2022error} was proposed, which requires two forward computations. First, network responses to the input and, second, to its perturbed version.
Then, the resulting approximate gradient was used to compute the synaptic updates.
The model trained using these local learning rules has been shown to produce competitive results to BP \citep{nokland2016direct,bartunov2018assessing, crafton2019direct}.

\hlMethod{Three-factor learning}
In this method \cite{fremaux2016neuromodulated,gerstner2018eligibility}, the concurrent activation of a pre-synaptic and post-synaptic neuron is recorded in an eligibility trace that decays over time and marks the weights eligible for being updated. However, updating the weights requires an additional neuromodulatory signal which is the third factor. Eligibility Propagation (e-prop, \cite{bellec2020solution}) is a successful variant that can be applied to training spiking neural networks, but still learns slower than BPTT.

\hlMethod{Weight Perturbation (WP) Learning}  
The Finite Difference Stochastic Approximation (FDSA) method \cite{kiefer1952stochastic}, the earliest known gradient approximation algorithm using WP, followed by  \cite{barto1987gradient, mazzoni1991more, williams1992simple} generates a global reinforcement signal that can be broadcasted to hidden synapses or neurons. These methods involve perturbing a single weight \cite{jabri1991weight} or a neuron \cite{widrow199030}. Such one-at-a-time approaches suffer from time complexity that scales linearly with the number of weights $\mathcal{O}(W)$ and neurons $\mathcal{O}(N)$ per iteration $t\in T$ to approximate the gradient. 
Later, \citet{dembo1990model} proposed model-free distributed learning with parallel WP to tackle slow training, followed by \citet{salimans2017evolution} proposing simultaneous perturbation stochastic approximation with learning rates modulated by Adam. 
Recently, \citet{zuge2023weight} explored similar ideas with simultaneous WP \citep{alspector1992parallel}, claiming its superiority over node perturbation on temporally extended time series forecasting benchmarks.
Further, the resulting gradient approximation methods have a close resemblance to Follow The Perturbed Leader \citep[FTPL,][]{kalai2005efficient, syrgkanis2015fast} optimization frameworks. Such methods use a proxy error function called regret score, which measures the difference between the expected outcome from the leader and the actual outcome from the follower network.

Although, these methods have addressed several limitations of gradient-based BP,
they still struggle in two key domains, biological realism, and/or computational efficiency \cite{jaderberg2017decoupled,bengio1994learning}.
Here we extend WP with an adaptive learning rate to develop a biologically more realistic learning rule and improve the convergence speed and stability compared to the previous methods.

\begin{figure*}[!t]
    
    \begin{subfigure}[b]{0.32\textwidth}
         \centering
         \caption{}
         \includegraphics[width=\linewidth]{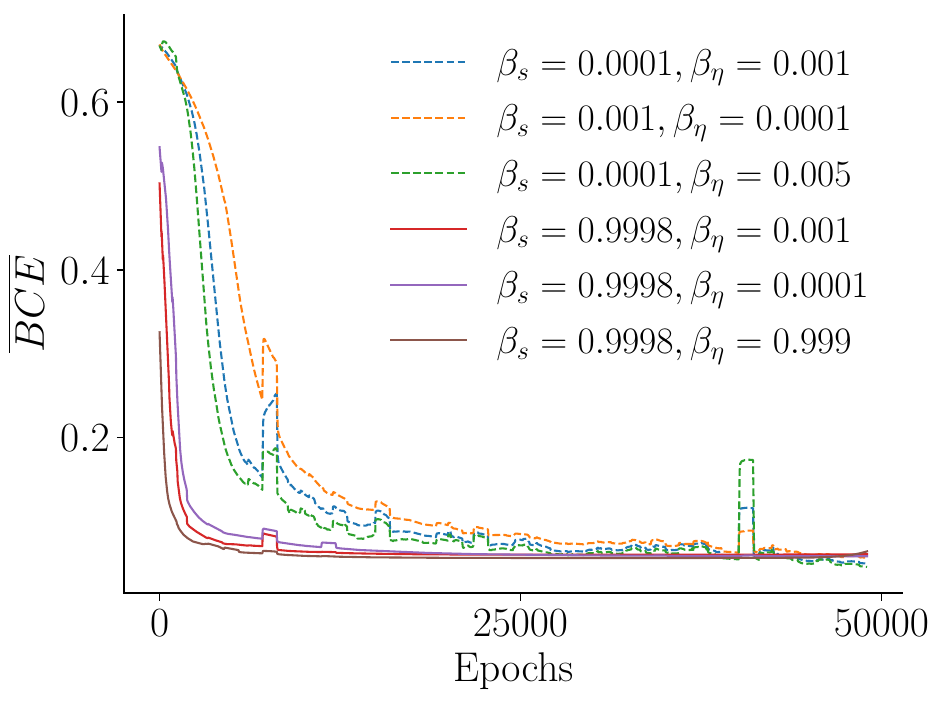}  
    \end{subfigure}
	\begin{subfigure}[b]{0.32\textwidth}
		\centering
        \caption{}
		\captionsetup[subfloat]{labelformat=empty}
		\includegraphics[width=\textwidth, height=0.7\textwidth]{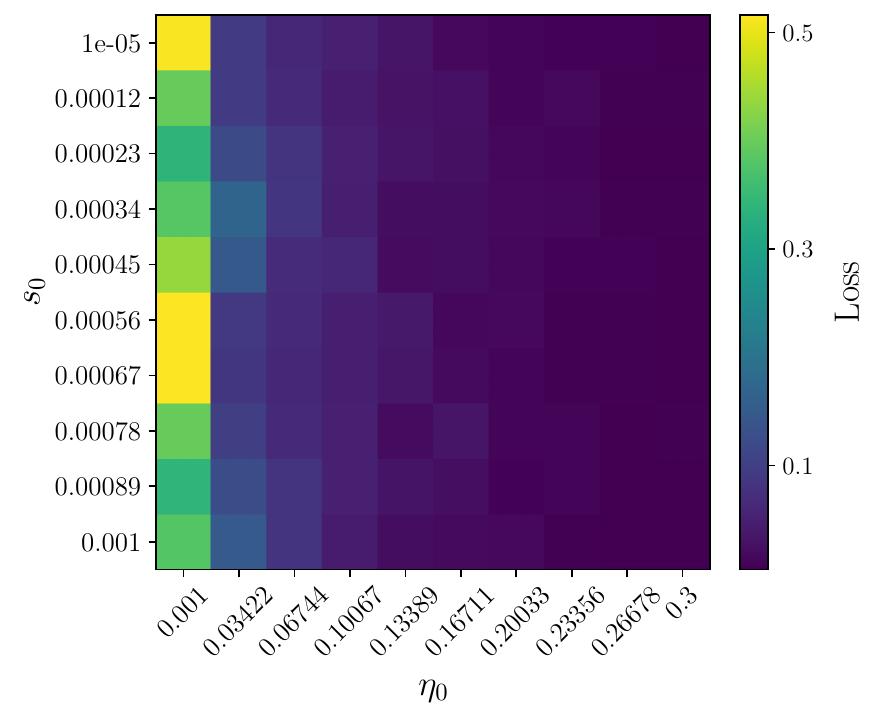}
    \end{subfigure}
    \begin{subfigure}[b]{0.32\textwidth}
		\centering
		\caption{}
	  \includegraphics[width=\textwidth, height=0.7\textwidth]{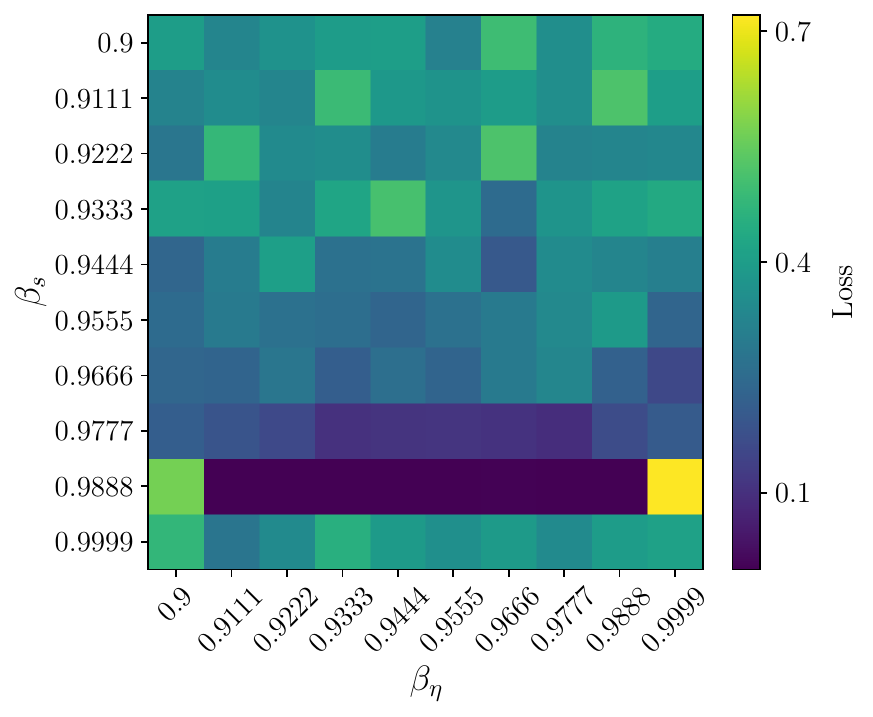}
	\end{subfigure}
    \caption{Impact of hyperparameters in Dopamine algorithms. a) Training curves for Dopamine 1 (dotted) and Dopamine 2(solid) for different moving average coefficient values ($\beta_{s},\beta_{\eta}$). See \autoref{tab:xor_hyperparam_ablation} for detailed hyperparameter settings. b) Influence of initial value of auxiliary variable $s_0$ and initial learning rate $\eta_0$ on the final loss for an RNN trained on Rossler attractor data. c) Impact of moving average coefficients on the performance of the RNN trained on Rossler attractor data. Detailed hyperparameter settings are provided in \autoref{tab:rossler_hyperparam_ablation}.
    \vspace{-10pt}
    }
    \label{fig:betas_etas}
\end{figure*}

\section{Dopamine's update rule}\label{sec:DopamineRule}

Inspired by the key role of the brain's dopaminergic system and the previous approaches \citep[in particular weight perturbation learning, ][]{jabri1991weight,jabri1992weight,cauwenberghs1992fast,zuge2023weight}, we introduce \emph{Dopamine's update rule}.
Several pieces of empirical neurobiological evidence support that dopaminergic neurons measure discrepancies between the expected and actual outcome, widely recognized as Reward Prediction Error \citep[RPE,][]{schultz1997neural, bush1951mathematical,rescorla1972theory,watabe2017neural}. Further, the Basic Synaptic Noise Hypothesis (BSNH) implies that noise from various biological synaptic sources \citep{steinmetz2000subthreshold} influences neural network learning \citep{korn1986probabilistic}.
Thus, based on the fundamental idea of RPE and BSNH, we propose a derivative-free and computationally efficient optimizer for training neural networks. 
Specifically, we use RPE to adjust the learning rate in the network, creating an adaptive learning rate strategy, similar to the role of dopamine in the brain.
In the following, we provide the mathematical formulation of our optimizer for training neural networks.

Given a dataset $D=\{(x_t,y_t)_{t=1}^T\}$ where $x_t \in \mathbb{R}^k$ is the $t$-th input feature vector, $y_t \in \mathbb{R}$ is the corresponding scalar target and $\theta$ represents parameters of the mapping function, $f \colon x \rightarrow y$,
which is Lipschitz continuous, the objective is to solve the following minimization problem,
\begin{equation}
\label{eqn:empiricalrisk}
    \min_{\theta \in \mathbb{R}^d} f(\theta):= \frac{1}{T} \sum_{t=1}^{T}\mathcal{L}(\theta,x_t)\,, 
\end{equation}
where $\mathcal{L}$ is a (convex) loss function. Let us assume, there exists a leader $f^*$ with optimal parameter $\theta^*$ in hindsight similar to FTPL \citep{kalai2002geometric, kalai2005efficient,lee2024follow}. We then formulate the above empirical risk minimization problem \autoref{eqn:empiricalrisk} as a regret optimization problem \citep{mahdavi2012efficient, jin2018regret}. Under online learning settings \citep{cohen2015following}, the objective is to minimize the regret \autoref{eqn:regretff}, the difference in cumulative sum of the loss incurred by the learner to the loss incurred by the leader: $\theta^*= \theta+\xi$; where $\xi \in \mathbb{R}^d$ is distributed over $\mathcal{N}(0,\sigma^{2} I)$,
 \begin{equation}
	\label{eqn:regretff}
	\resizebox{.9\hsize}{!}{$\mathcal{R}(T) = \mathop{{}\mathbb{E}}_{\xi \sim \mathcal{N}(0,\sigma^2I)}\left[\sum_{t=1}^{T}\left(\hat{\mathcal{L}}(x_t,\theta,\xi) - \mathcal{L}(x_t,\theta)\right)\right]\,$}
\end{equation}
Hence, when $\sigma \in 0< \sigma <<1$ is small, we expect the loss incurred by the leader($\hat{\mathcal{L}}$) to be closer to the unperturbed loss $\mathcal{L}$ incurred by the learner and vice versa when $ \sigma \in 0<< \sigma <1$ is large. Thus, intuitively, \autoref{eqn:regretff} links the biological concept of RPE influenced by synaptic noise to the regret optimization problem using weight perturbation. Furthermore, as we discuss in the next section, we use the RPE or the regret (\autoref{eqn:regretff}) function introduced above to adapt the learning rate $\eta$.

Lastly, the discussed optimization framework optimizes the global objective function $\mathcal{L}$ using the following parameter update rule,
\begin{equation}
    \label{eqn:weightupdate}
   \Delta \theta = - \frac{\eta}{\sigma^2} \mathcal{R}\xi\,.
\end{equation}
%
In contrast to previous studies \cite{jabri1992weight}, the components of the global gradient are obtained using \emph{simultaneous perturbation method}, for instance, across all the layers in the network \cite{cauwenberghs1992fast,zuge2023weight}, instead of sequentially perturbing single weight or node at a time. Therefore, when $\mathcal{R}(T)$ is positive, the perturbation $\xi$ has positive components towards the reward gradient, and the parameters are updated along the direction of the perturbation. Conversely, when $\mathcal{R}(T)$ is negative, the parameter update happens against the direction of the perturbation. Ultimately, moving against the direction of the global gradient iteratively minimizes the loss $\mathcal{L}$.

\begin{figure}[!t]
    \begin{minipage}{0.62\textwidth} 
        \begin{subfigure}[b]{0.4\textwidth}
            \captionsetup[subfigure]{position=top}
            \caption*{BP-Adam}
            \includegraphics[width=\textwidth, height=0.7\textwidth]{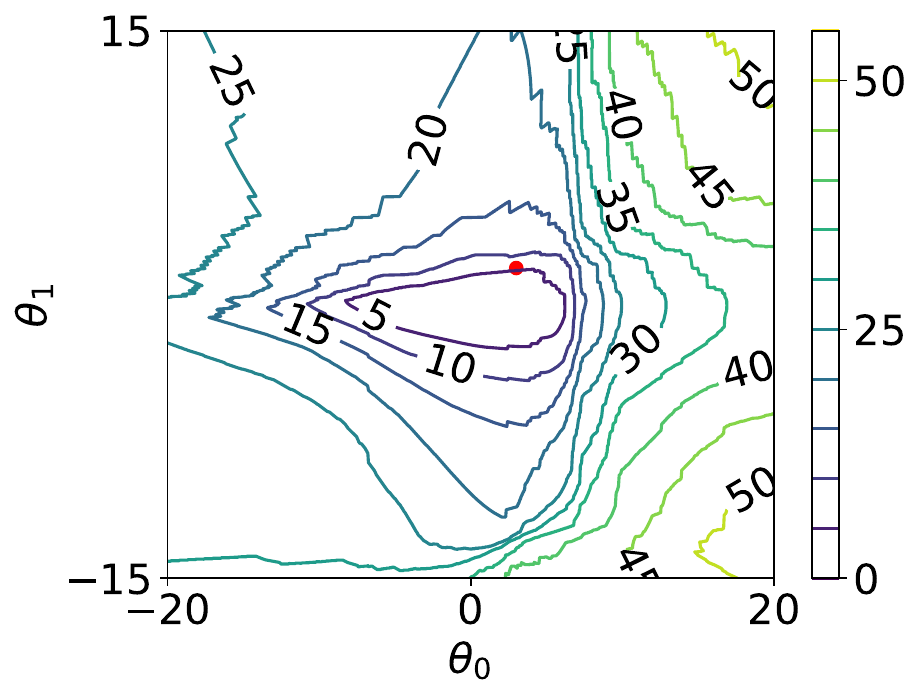}
        \end{subfigure}
        \hspace{-0.5em} 
        \begin{subfigure}[b]{0.4\textwidth}
            \caption*{Dopamine 2}
            \includegraphics[width=\textwidth, height=0.7\textwidth]{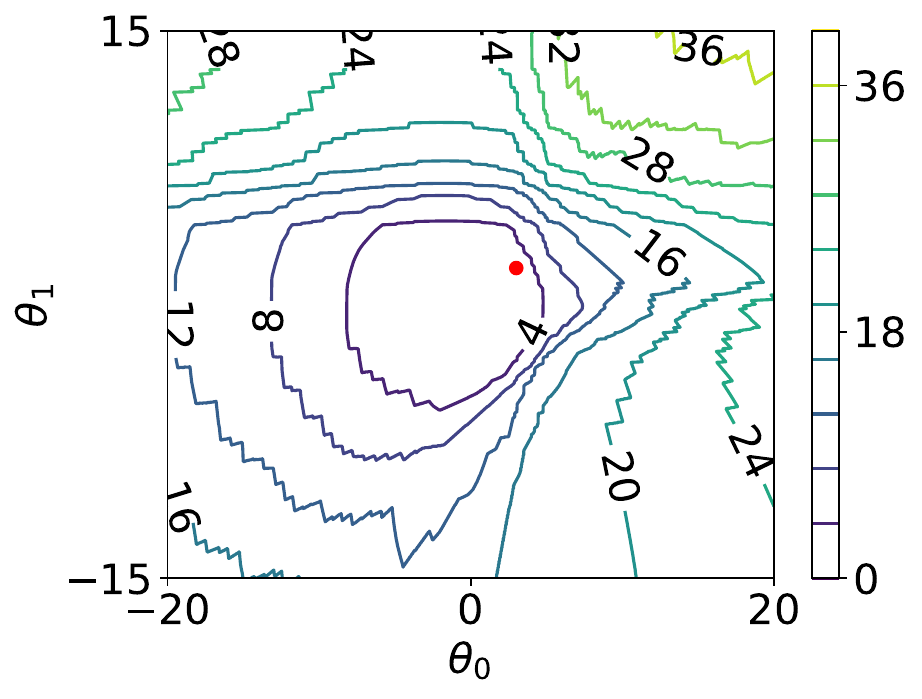}
        \end{subfigure}
        
        \begin{subfigure}[b]{0.4\textwidth}
            \includegraphics[width=\textwidth, height=0.7\textwidth]{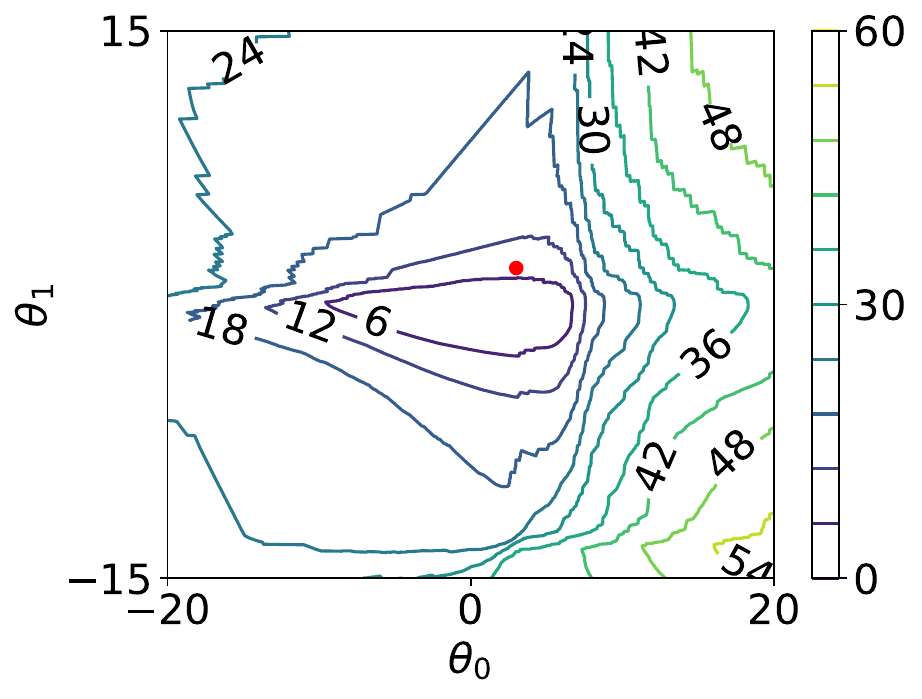}
        \end{subfigure}
        \hspace{-0.5em}
        \begin{subfigure}[b]{0.4\textwidth}
            \includegraphics[width=\textwidth, height=0.7\textwidth]{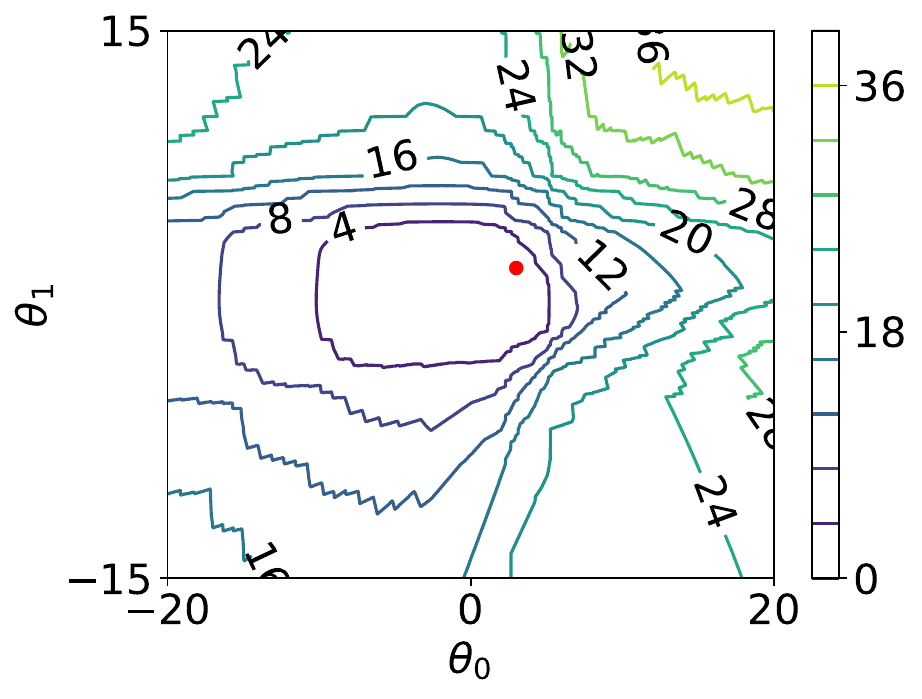}
        \end{subfigure}
    \end{minipage}%
    \hspace{-5em}
    \begin{minipage}{0.48\textwidth}
    \caption{XOR classification task: Loss Landscape computed by random direction method \citep{li2018visualizing} using 1000 steps in [-20, 20] for $\theta_0$ and [-15, 15] for $\theta_1$. The resulting loss (color) surface discretized into 10 contour regions. 
    Red dot indicates the coordinates of the optimal parameters on the loss landscape. 
     See \autoref{tab:xor_hyperparam} for hyperparameter details and \autoref{fig:loss_landscape} for other benchmarks.
     \vspace{-20pt}
     }
    \label{fig:loss_landscape_mainText}
    \end{minipage}
\end{figure}

\subsection{Adaptive learning rule}\label{"dopamine"}
Following recent developments in perturbation-based neural network optimization \citep{zuge2023weight,salimans2017evolution}, we introduce an adaptive learning rate (for $\eta$ in \autoref{eqn:weightupdate}).
To build the adaptive learning rate, and also to preserve the dynamics of the recurrent neural networks within the stable regime, we take the following steps
\footnote{It should be noted that, preserving the dynamics of the recurrent neural networks within the stable regime is crucial when the training happens under random perturbation.}. 
Firstly, we derive a biologically plausible parameter update rule with an adaptive learning rate strategy for simultaneous weight perturbation learning; secondly, we introduce a necessary constraint in the optimization process to preserve the dynamics of the RNN when influenced by random perturbation.

\begin{figure*}[!t]
    \centering
    \begin{minipage}{\textwidth} 
        \centering 
        \begin{subfigure}[b]{0.2\textwidth}
            \captionsetup[subfigure]{position=top}
            \caption*{BP-Adam}
            \includegraphics[width=\textwidth, height=0.7\textwidth]{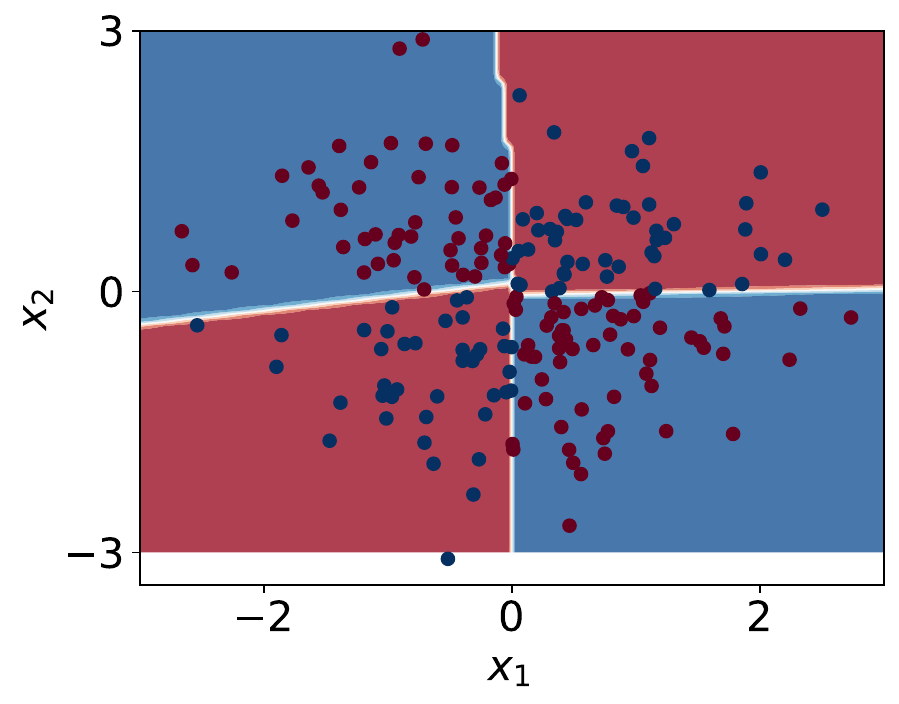}
        \end{subfigure}
        \hspace{-0.5em} 
        \begin{subfigure}[b]{0.2\textwidth}
            \caption*{DNI-Adam}
            \includegraphics[width=\textwidth, height=0.7\textwidth]{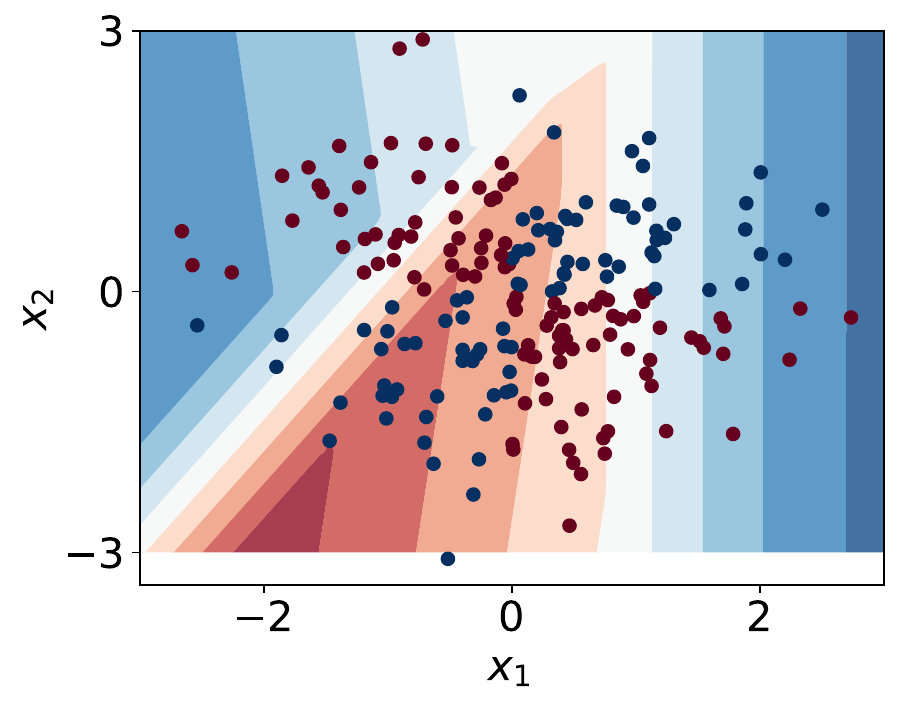}
        \end{subfigure}
        \hspace{-0.5em}
        \begin{subfigure}[b]{0.2\textwidth}
            \caption*{Dopamine-$1$}
            \includegraphics[width=\textwidth, height=0.7\textwidth]{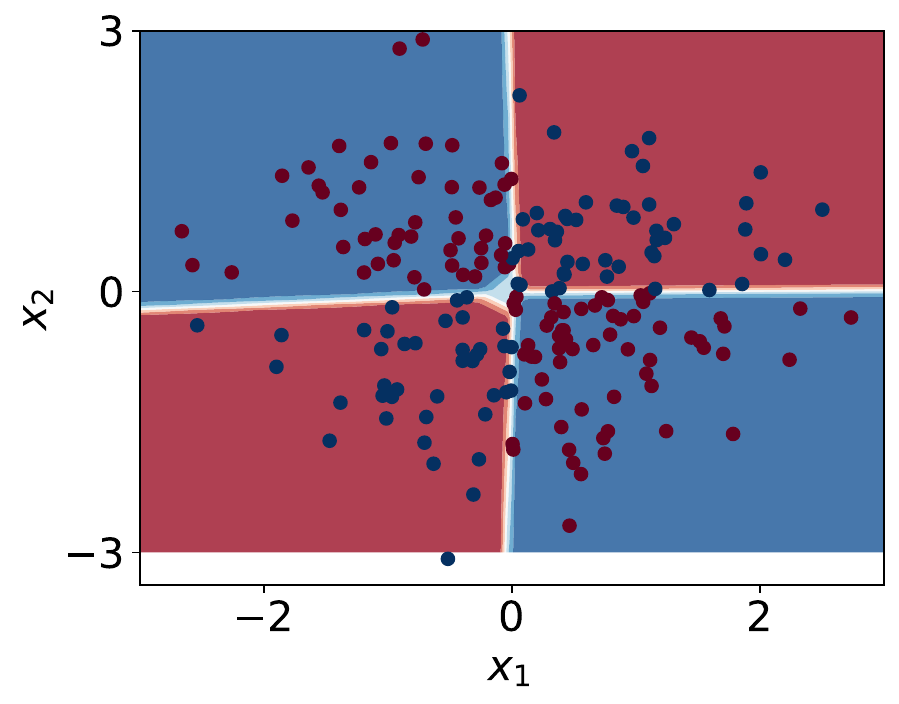}
        \end{subfigure}
        \hspace{-0.5em}
        \begin{subfigure}[b]{0.2\textwidth}
            \caption*{Dopamine-$2$}
            \includegraphics[width=\textwidth, height=0.7\textwidth]{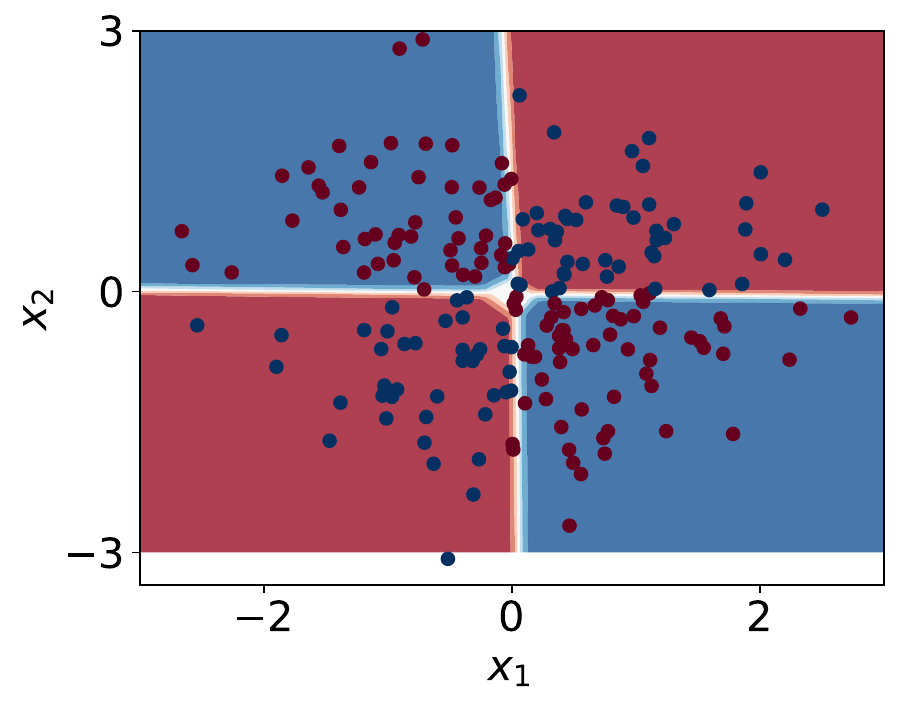}
        \end{subfigure}
        \\ 
        \begin{subfigure}[b]{0.2\textwidth}
            \includegraphics[width=\textwidth, height=0.7\textwidth]{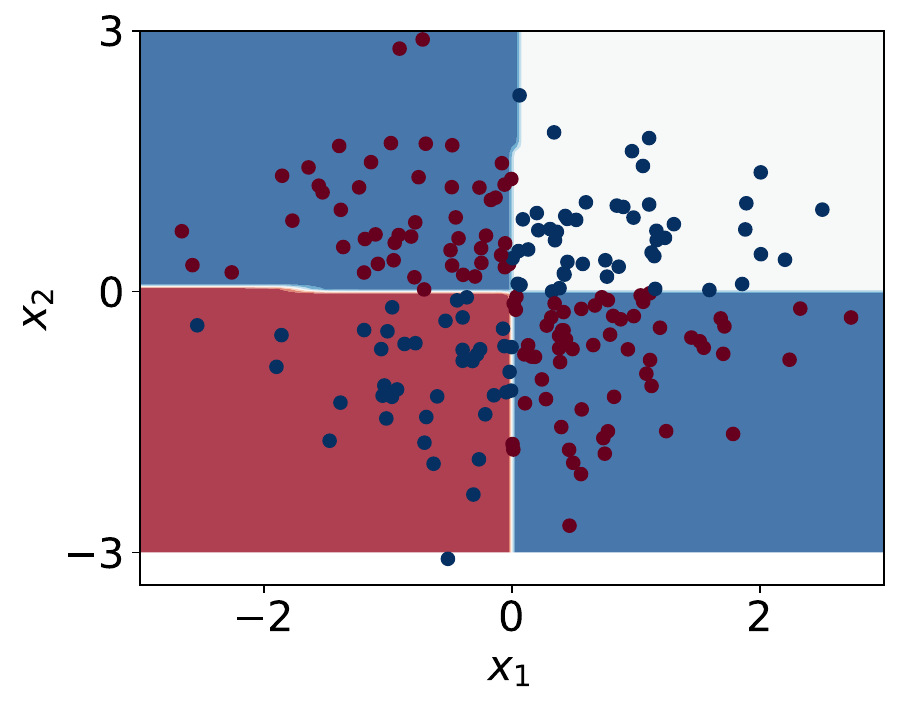}
        \end{subfigure}
        \hspace{-0.5em}
        \begin{subfigure}[b]{0.2\textwidth}
            \includegraphics[width=\textwidth, height=0.7\textwidth]{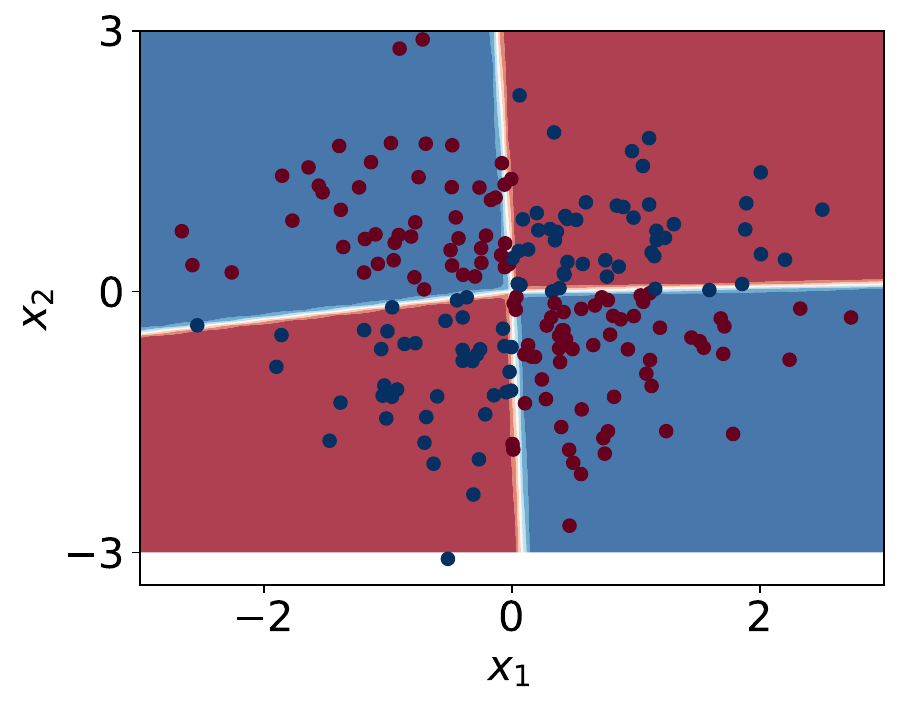}
        \end{subfigure}
        \hspace{-0.5em}
        \begin{subfigure}[b]{0.2\textwidth}
            \includegraphics[width=\textwidth, height=0.7\textwidth]{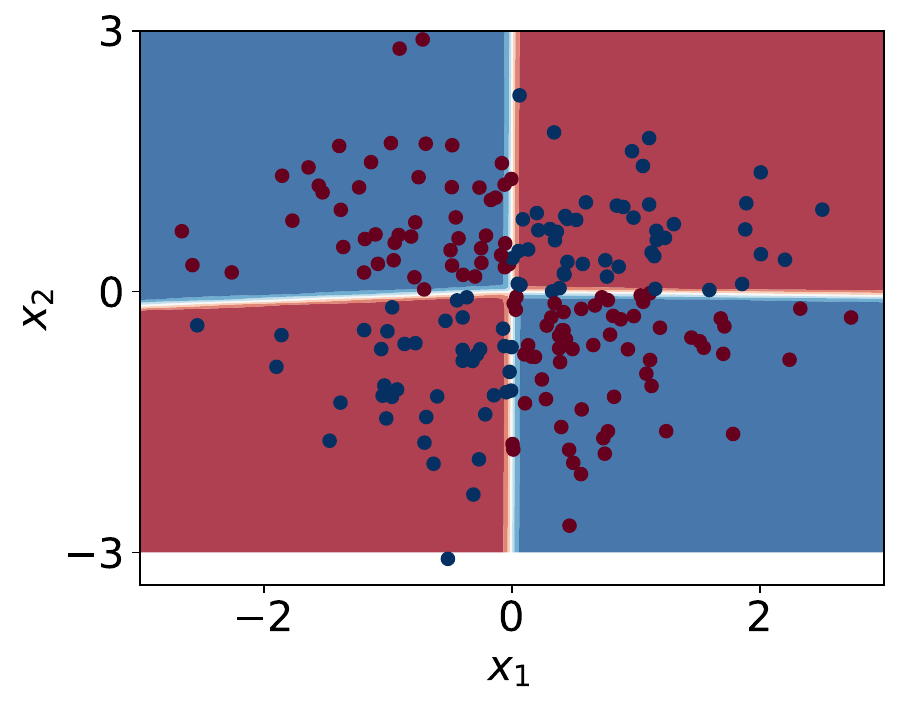}
        \end{subfigure}
        \hspace{-0.5em}
        \begin{subfigure}[b]{0.2\textwidth}
            \includegraphics[width=\textwidth, height=0.7\textwidth]{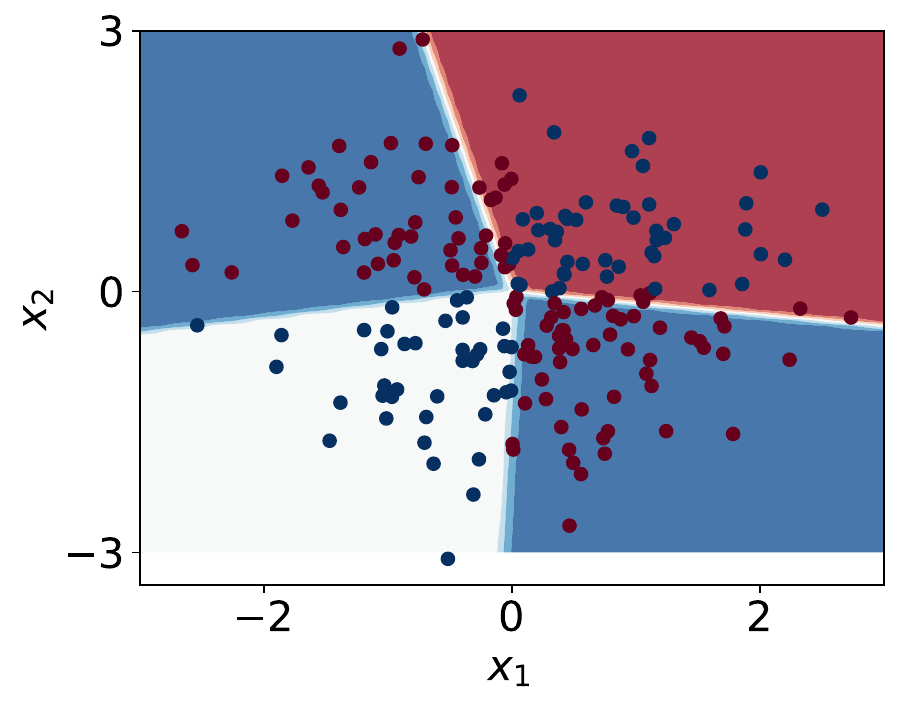}
        \end{subfigure}
    \end{minipage}
    \caption{Decision boundary of models on XOR classification task. Top row: Models with affine layers. Bottom row: Models with linear layers. BP-Adam refers to Back-Propagation with Adam. See \autoref{tab:xor_hyperparam} for hyperparameter details.}
    \label{fig:decision_boundary}
    \vspace{-1em}
\end{figure*}

We propose that learning rate should smoothly follow the RPE or regret ($\mathcal{R}$).
The computational rationale behind this choice is that, when the prediction of the learner is small (i.e., $\mathcal{R}$ in \autoref{eqn:regretff} has a small value), a proportionally small adjustment in learning model parameters is needed (thus, a small learning rate $\eta$ is needed). Conversely, when the learner's prediction is large, it implies that the learner's model parameters were not good enough for the changes (that was induced by the perturbation). Thus, a proportionally more significant adjustment (higher learning rate, $\eta$) is needed.
We mathematically formulate this postulate as follows (a heuristic that learning rate \emph{smoothly} tracks the reward gradient).
At the training step $t$, the learning rate $\eta$ in \autoref{eqn:weightupdate} is updated using the moving average of the reward gradient $\mathcal{R}$. 
With constant coefficient $\beta_s$, intermediate variable $s_t$ (\autoref{eqn:s_stochastic}) is simply the low (when $0<\beta_s<1$) pass filtered $\mathcal{R}$ (see Figure \autoref{fig:dopamine_update_rules}),
\begin{equation}
	\label{eqn:s_stochastic}
	s_t = \beta_{s}s_{t-1} - (1- \beta_s){\mathcal{R}_t} \ \ \text{where} \ \ 0<<\beta_s<1\,,
\end{equation}
\begin{equation}
	\label{eqn:eta_follow_reward}
        \resizebox{.9\hsize}{!}{$\eta_t = (1-\beta_{\eta})\eta_{t-1} - \beta_\eta s_t \ \ \text{where} \ \ 0<\beta_{\eta}<1 \ \ \text{\&} \ \ \beta_{\eta} < \beta_{s}\,.$}
\end{equation}
Thus, $\beta_s$ and $(1-\beta_\eta)$ act as decay factors of the learning rate. Hence, the above proposed adaptive learning rate method updates $\eta$ proportional to the moving average of $\mathcal{R}$ under constant coefficients $\beta_s$ and $\beta_{\eta}$ (\autoref{fig:dopamine_update_rules}). Such update rule, where $\eta$ is proportional to the reward signal, is based on the modest assumptions based on the biological evidence observed during Reinforcement Learning \citep{urbanczik2014learning,coddington2023mesolimbic}. Hence, we call it \textit{Dopamine-1}. Though successful in training feedforward networks on classification tasks (\autoref{fig:decision_boundary}), there is a caveat. \textit{Dopamine-1}, \autoref{eqn:eta_follow_reward}, pushes the initial learning rate instantly towards $\mathcal{R}$, which is unknown. On the other hand, training RNNs demands fine-tuned initial hyperparameter settings, especially the learning rate. Hence, to address this caveat, we propose \textit{Dopamine-2},
\begin{equation}
	\label{eqn:eta_exponential_decay}
	\eta_t = (1-\beta_{\eta})\eta_{t-1} + \beta_\eta{s_t}\,,
\end{equation} 
that exponentially decays the learning rate towards the reward, see \autoref{fig:dopamine_update_rules}.

Hence, the parameter update rule given $\eta_t$ at training iteration $t$ is as follows,
\begin{align}
	\label{eqn:grad_descent}
	\theta_{t+1} \leftarrow \theta_t - \eta_t \mathcal{R}_t \frac{\xi_t}{\sigma_t ^2}\,,
\end{align}
since $\langle \xi_t \rangle=0$,then  $\langle \partial\theta \rangle \approx - \langle \mathcal{R}_{\beta_\eta \beta_s
} \rangle \frac{\partial \mathcal{L}}{\partial \theta}$ and $\eta \approx \langle \mathcal{R}_{{\beta}_{\eta}\beta_{s}}\rangle $. Notably, the parameter update step for RNNs using truncated gradient approximation for an infinite length time series is,
\begin{align}
\label{eqn:rnndescent}
\theta_{t} \leftarrow \theta_{t} - \eta_{t}  
\underbrace{\left[\sum_{\tau=t}^{t+T} \Delta\mathcal{L}_{\tau}\right]}_{\mathcal{R}_\tau} 
\frac{\xi_t}{\sigma_{t}^{2}},
\end{align}
where  $\Delta\mathcal{L}_{\tau} = \hat{\mathcal{L}}_{\tau}(x_\tau, \theta_t, \xi_t) - \mathcal{L}_{\tau}(x_\tau, \theta_t)$ and $\tau$ represents truncated sequence length. However, the above parameter update expression is an ill-fated problem due to the stochastic nature of $\xi$ (especially on chaotic time series prediction tasks). Therefore, for guaranteed learning, an additional step is taken to maintain the RNN dynamics in a stable regime \cite{jaeger2002adaptive}. That is, at regular training intervals, we reset the spectral radius, perhaps the largest absolute eigenvalue of the recurrent weights $\theta^{rec}$ closer to 1, such that $1-|\epsilon| <\rho(\theta^{rec})< 1+|\epsilon|$, where $|\epsilon|$ represents a small positive real value(see \autoref{appx:spectral_radius}). We call it \textit{Spectral Weight Perturbation} (Spectral WP) learning. 


\begin{algorithm}[!h]
   \caption{Optimize RNNs using \textit{Dopamine}}
   \label{alg:dopamine}
\begin{algorithmic}
    
   \REQUIRE {\bfseries Input:} Learning rate $\eta$, perturbation standard deviation $\sigma$, exponential decay factors $\beta_{s}$ and $\beta_{\eta}$, spectral radius $\lambda$
   \REQUIRE {\bfseries Initialize:} Model parameters $\theta$, objective function $\mathcal{L}$
   \ENSURE {\bfseries Ensure:} $\rho(\theta^{rec}) = \lambda$ \COMMENT{Set spectral radius}
   \FOR{$t = 0, 1,2, \dots$}
   \FOR{$l = 0, 1, 2, \dots$} 
   \STATE $\xi^l \sim \mathcal{N}(0, \sigma^2I)$ \COMMENT{Sample perturbation noise for layer $l$}
   \ENDFOR
   \STATE $\mathcal{R}_t \leftarrow \hat{\mathcal{L}}(\theta_t, \xi) - \mathcal{L}(\theta_t)$ \COMMENT{Global reward gradient}
   \FOR{$l = 0, 1, 2, \dots$} 
   \STATE $s_t \leftarrow \beta_{s}s_{t-1} - (1- \beta_s){\mathcal{R}_t}$  
   
   \STATE $\eta_t^{l} \leftarrow (1-\beta_{\eta})\eta_{t-1}^{l} + \beta_\eta{s_t}$ \COMMENT{Dopamine-2}
   
   \STATE $\theta_{t+1}^l \leftarrow \theta_{t}^l - \eta_{t}^l \mathcal{R}_t \frac{\xi^l}{\sigma_{l}^2}$
   
   \STATE $\rho(\theta^{rec}_{t+1}) \leftarrow \lambda$ \COMMENT{Reset spectral radius}
   \ENDFOR
   \ENDFOR
\end{algorithmic}
\end{algorithm}

 To understand the impact of the hyperparameters on model performance, we analyzed the influence of exponential moving average (EMA) coefficients $\beta_s$ \& $\beta_{\eta}$, the initial value of the auxiliary variable $s_0$ and initial learning rate $\eta_0$. The analysis was conducted on two tasks: Feed- Forward Neural Network for XOR classification(\autoref{fig:betas_etas}, \textit{Left}) and RNN for one step-ahead prediction using Rössler attractor dataset  (\autoref{fig:betas_etas}, \textit{Center}). Under varying initial conditions of the EMA, we found that the \textit{Dopamine 2} trained models outperformed those trained with \textit{Dopamine 1} on classification task. Specifically, Dopamine 2 showed convergence to better minima when the $s_0< \eta_0$ and $\beta_{s}>\beta_{\eta}$. 
%

\section{Experiments}\label{sec:experiments}
\subsection{Multi-Layered Perceptron: Classification tasks}

We compared our derivative-free optimization method with several other competing training algorithms that exploit true gradients (Adam) and approximate gradients (DNI-Adam) on a classical XOR classification problem. We used two types of fully connected Feed-Forward Neural Networks (FFNN); one with affine layers and another with linear layers. Each of these networks has a single hidden layer of size 4. The hidden layer uses Rectified Linear Unit (ReLU) and the output layer has sigmoid nonlinearity followed by \textit{softmax} function.  The networks were trained for 50k epochs with Binary Cross Entropy (BCE) as an objective using the competing optimizers (for details on hyperparameters, see, \autoref{tab:xor_hyperparam}).  We found that \textit{Dopamine-1} trained models were able to overcome constraints on the network layer's schema and learn the offset in the data without the bias term \autoref{fig:decision_boundary}. Furthermore, to test the robustness and generalization gap of the competing optimizers, we generated the loss surface of the trained models using the random direction perturbation method \citep{li2018visualizing}. We found that models trained using gradient-based optimizers converge to the saddle points (\autoref{fig:loss_landscape_mainText}, location of the red dot on the loss surface), whereas models trained using \textit{ dopamine} avoided the saddle points and converged more strongly inside the loss valley, leading to a better (lower) generalization gap between training and test data (\autoref{fig:loss_landscape_mainText} and \autoref{fig:loss_landscape}).
%
%
%
%
Further analytical studies on the optimal initial condition of the hyperparameters such as $s_0$ and the perturbation scale $\sigma$ could additionally improve the overall performance of the \textit{Dopamine} trained models. 
   \begin{figure*}[!t]
	\label{fig:chaotic_timeseries}
	\centering
	\begin{subfigure}[b]{0.24\textwidth}
		\captionsetup[subfigure]{position=top}
		\centering
		\caption*{BP-SGD}
 		\includegraphics[width=\textwidth, height=0.7\textwidth]{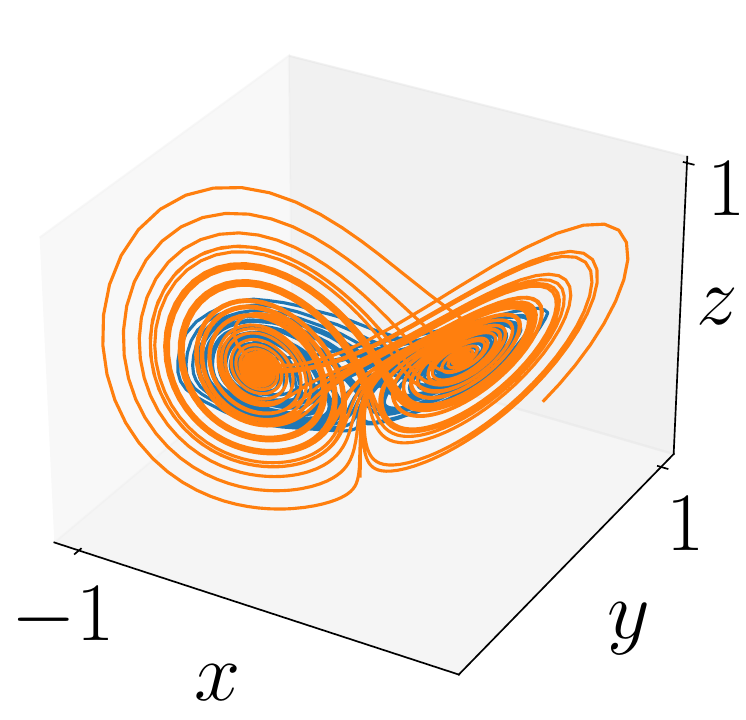}
	\end{subfigure}
	\hfill
	\begin{subfigure}[b]{0.24\textwidth}
		\centering
		\caption*{WP}
		\includegraphics[width=\textwidth, height=0.7\textwidth]{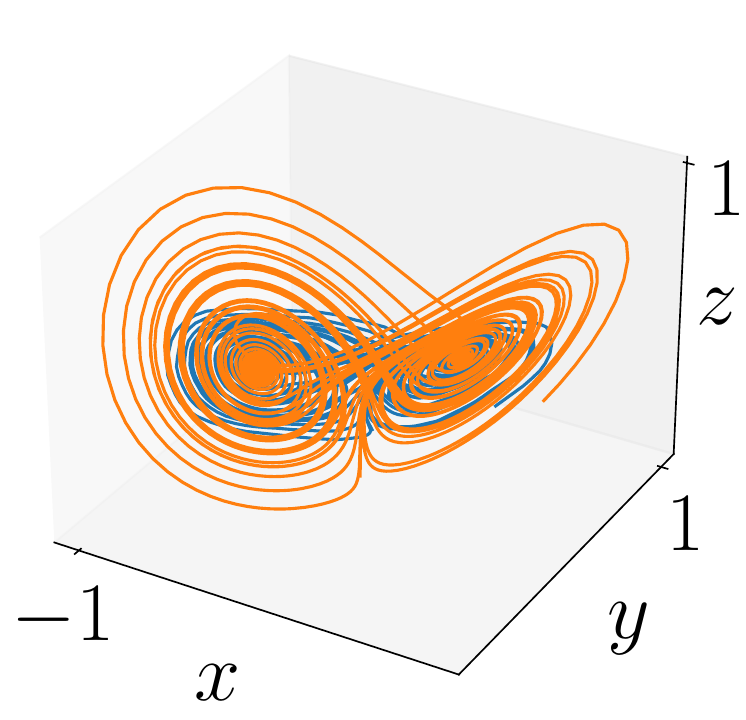}
	\end{subfigure}
	\hfill
	\begin{subfigure}[b]{0.24\textwidth}
		\centering
		\caption*{Spectral WP}
		\includegraphics[width=\textwidth, height=0.7\textwidth]{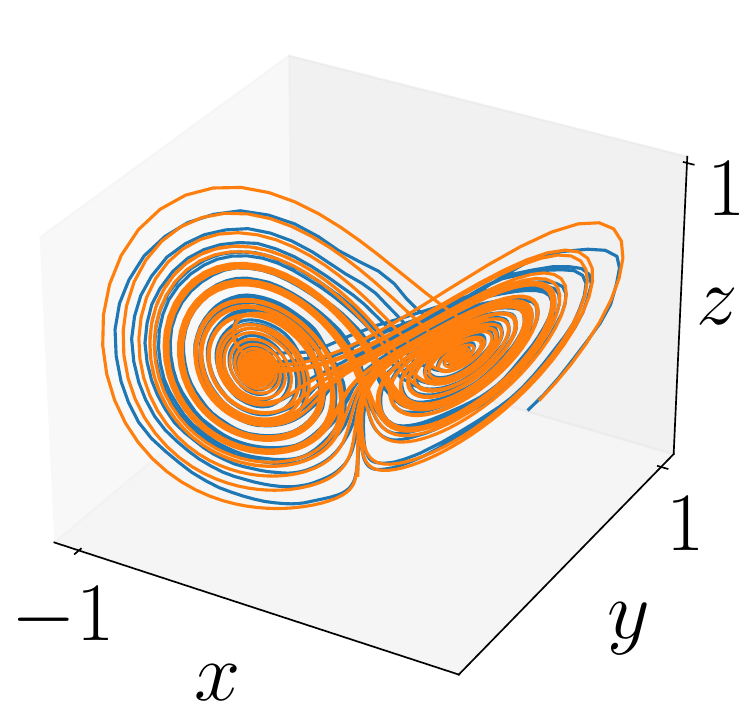}
	\end{subfigure}
	\hfill
	\begin{subfigure}[b]{0.24\textwidth}
		\centering
		\caption*{Dopamine-2}
	  \includegraphics[width=\textwidth, height=0.7\textwidth]{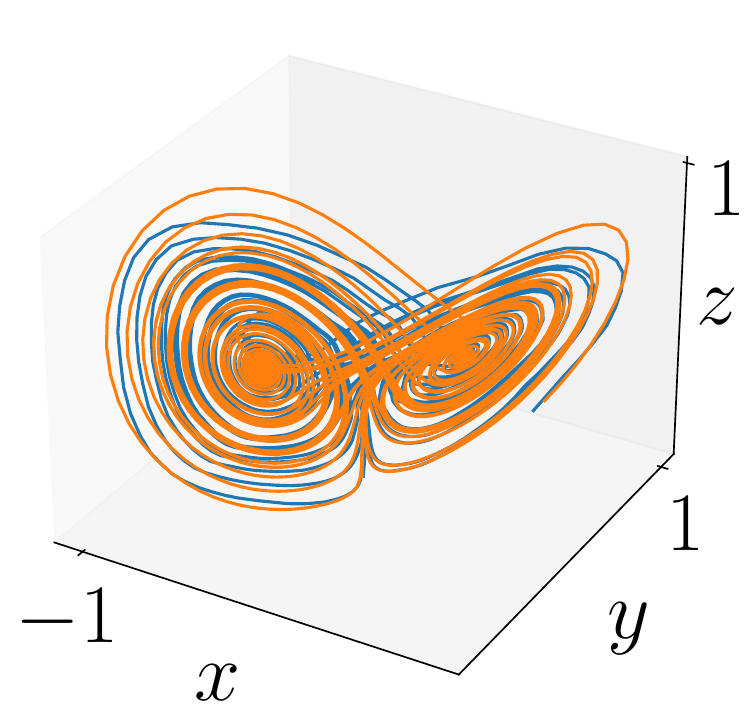}
	\end{subfigure}
	\hfill
	\begin{subfigure}[b]{0.24\textwidth}
		\centering   
		\includegraphics[width=\textwidth, height=0.7\textwidth]{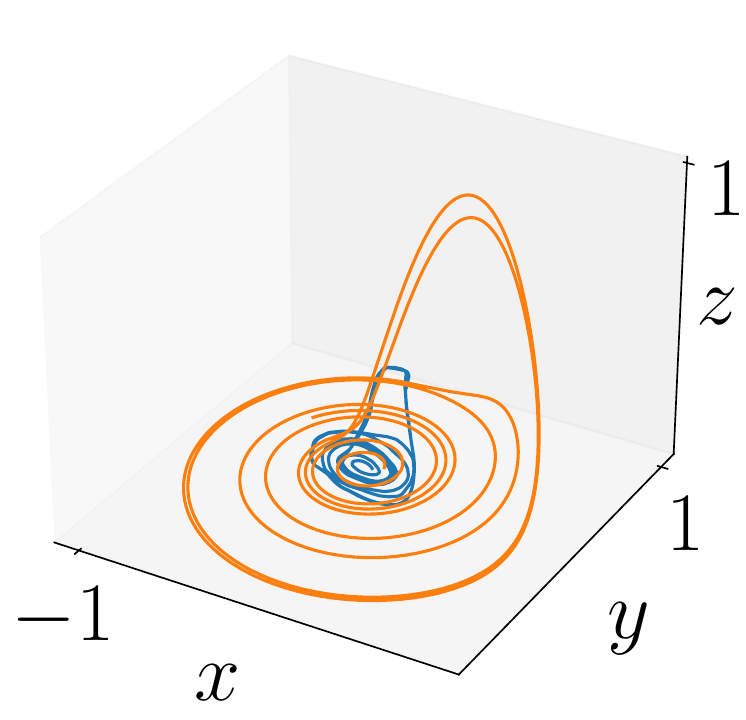}
	\end{subfigure}
	\hfill
	\begin{subfigure}[b]{0.24\textwidth}
		\centering
		\includegraphics[width=\textwidth, height=0.7\textwidth]{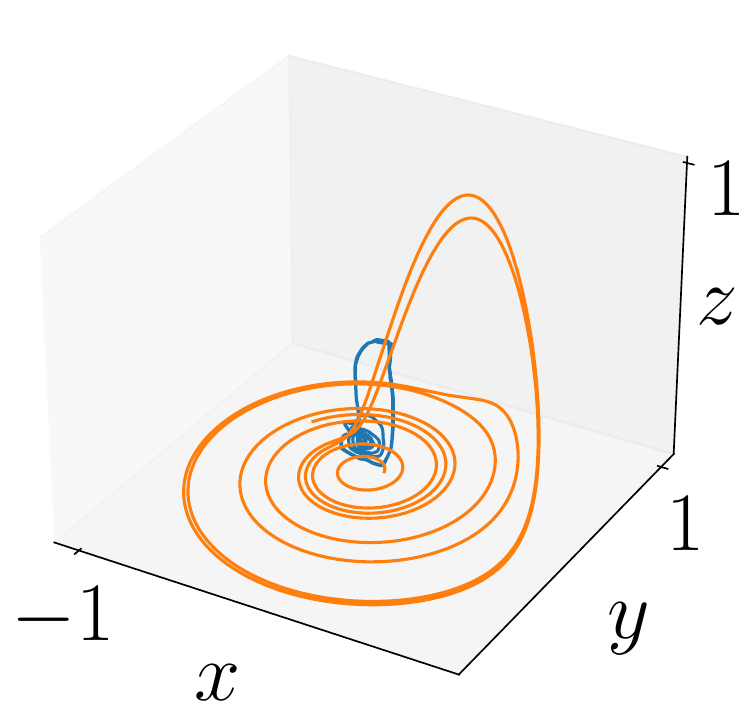}
	\end{subfigure}
	\hfill
	\begin{subfigure}[b]{0.24\textwidth}
		\centering
		\includegraphics[width=\textwidth, height=0.7\textwidth]{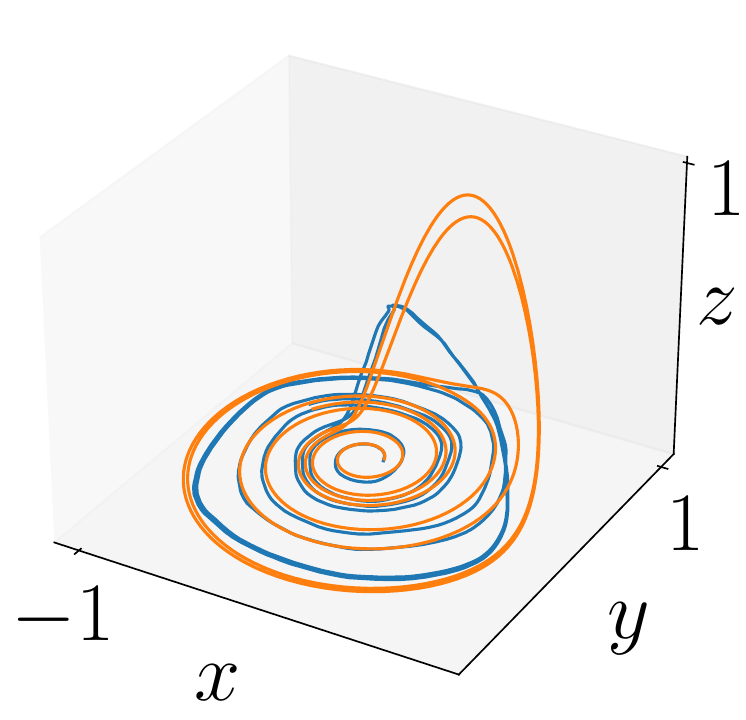}
	\end{subfigure}
	\hfill
	\begin{subfigure}[b]{0.24\textwidth}
		\centering
		\captionsetup[subfloat]{labelformat=empty}
		\includegraphics[width=\textwidth, height=0.7\textwidth]{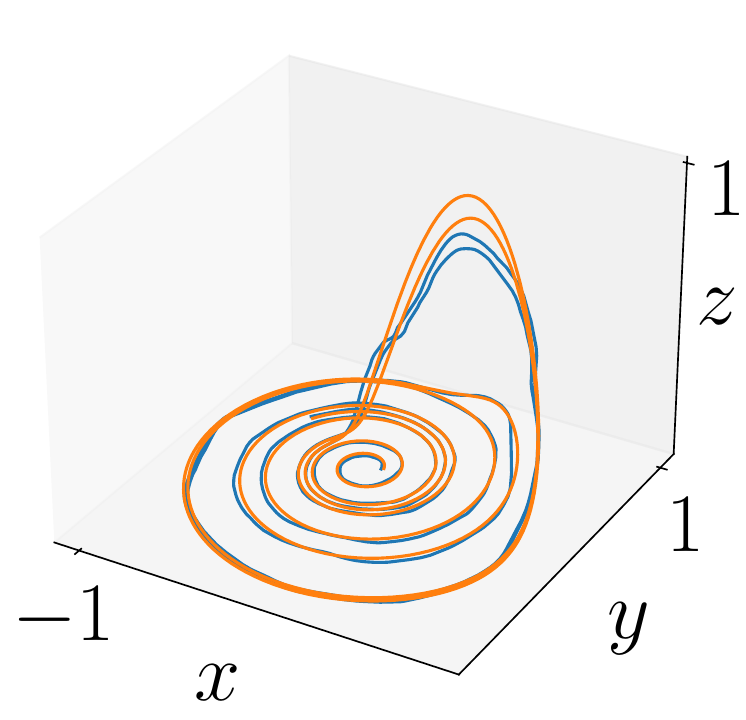}
	
	\end{subfigure}
	\hfill
        \vspace{0.3cm}
	\caption{\textit{Top row}: One step-ahead prediction on Lorenz attractor. \textit{Bottom row}: One step-ahead prediction on Rössler attractor. BP-SGD refers to Back-Propagation Through Time (BPTT) with Stochastic Gradient Descent. Titles indicate different optimizers. The orange traces are the original attractor and the blue traces are the reconstruction. See \autoref{tab:lorenz_hyperparam} and \autoref{tab:rossler_hyperparam} for hyperparameter details and \autoref{fig:adaptive_lr_lorenz_rossler} for the comparison of Adam and DNI-Adam against \textit{Dopamine 2}.
    \vspace{-10pt}
    }
    \label{fig:lorenz_rossler_predictions}
\end{figure*}

\begin{figure*}[!t]
    \centering
    \begin{subfigure}[t]{0.32\textwidth}
        \centering
        \includegraphics[width=1.\linewidth]{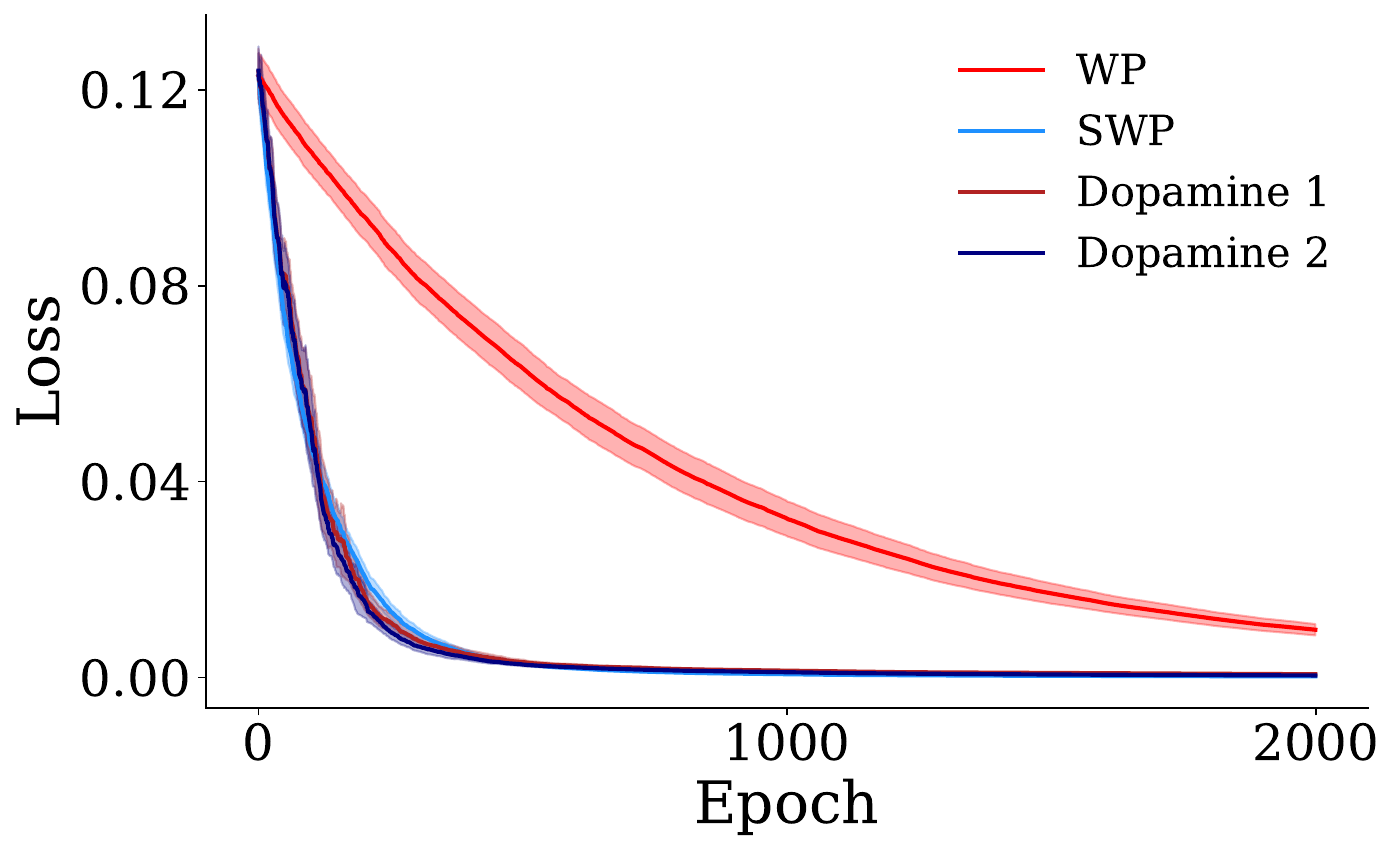}
        \label{fig:lorenz_loss_main}
    \end{subfigure}
    \begin{subfigure}[t]{0.32\textwidth}
        \centering
        \includegraphics[width=1.\linewidth]{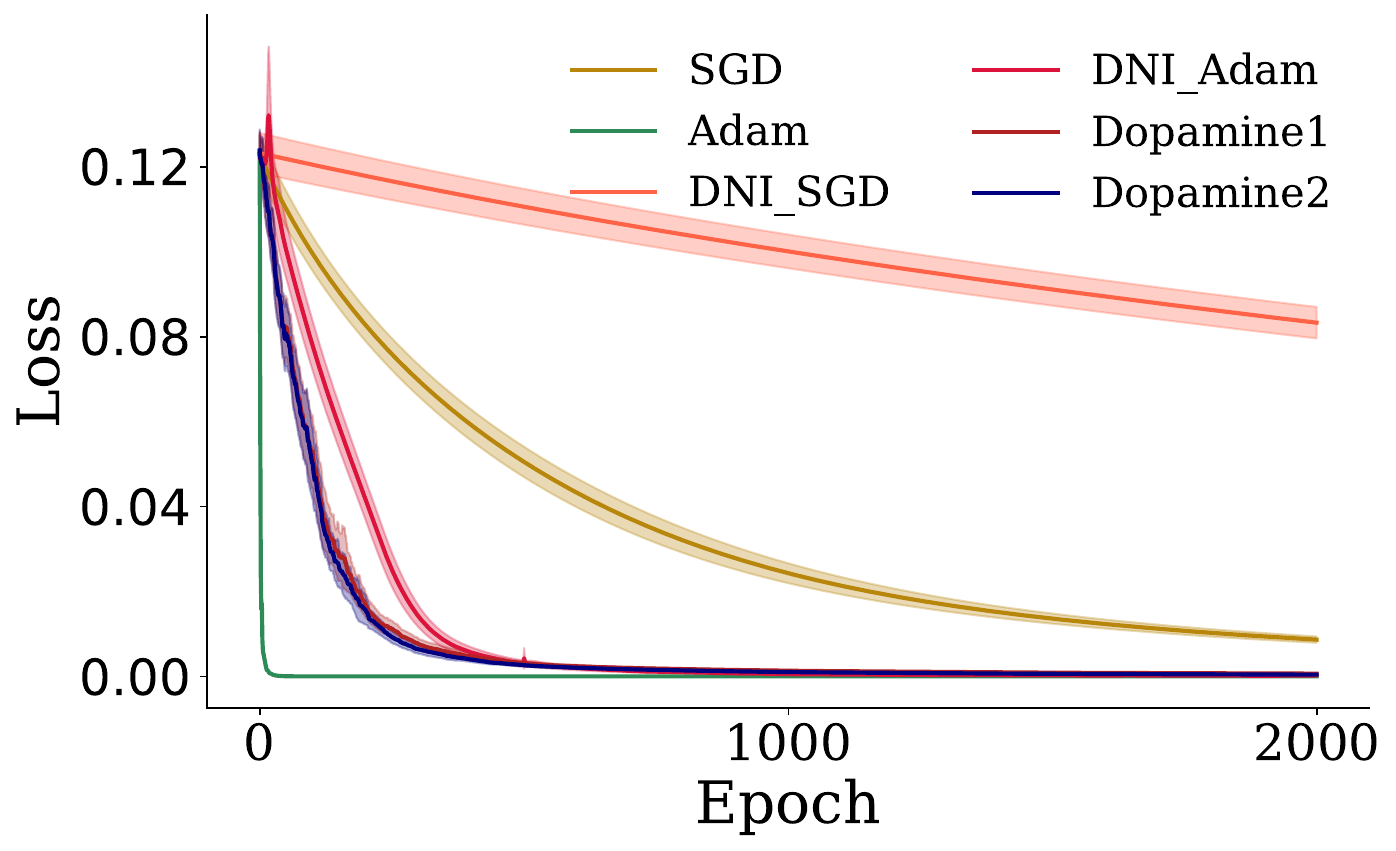}
        \label{fig:mnist_accs}
    \end{subfigure}
    \hfill{}\hspace{0.1\textwidth}
    \scalebox{0.75}{
    \begin{subfigure}[t]{0.32\textwidth}
        \centering
        \vspace{-12em}
        \hspace{-12em}
        \begin{tabular}{c|c} 
        \textbf{Lorenz} & Average final loss; (CI) \\ 
        \hline
        SGD & 0.0087; (0.0079, 0.0094) \\ 
        Adam & $1e^{-6}$; ($1e^{-6}$, $1e^{-6}$) \\ 
        DNI-SGD & 0.0833; (0.0796, 0.0871) \\ 
        DNI-Adam &  0.0002; (0.0001, 0.0004) \\ 
        WP & 0.0097; (0.0086, 0.0109) \\ 
        SWP & 0.0002; (0.0002, 0.0002) \\ 
        Dopamine1 & 0.0007; (0.0007, 0.0007) \\ 
        Dopamine2 & 0.0004; (0.0004, 0.0004) \\ 
        \end{tabular}
    \end{subfigure}}
    
    \begin{subfigure}[t]{0.32\textwidth}
        \centering
        \includegraphics[width=1.\linewidth]{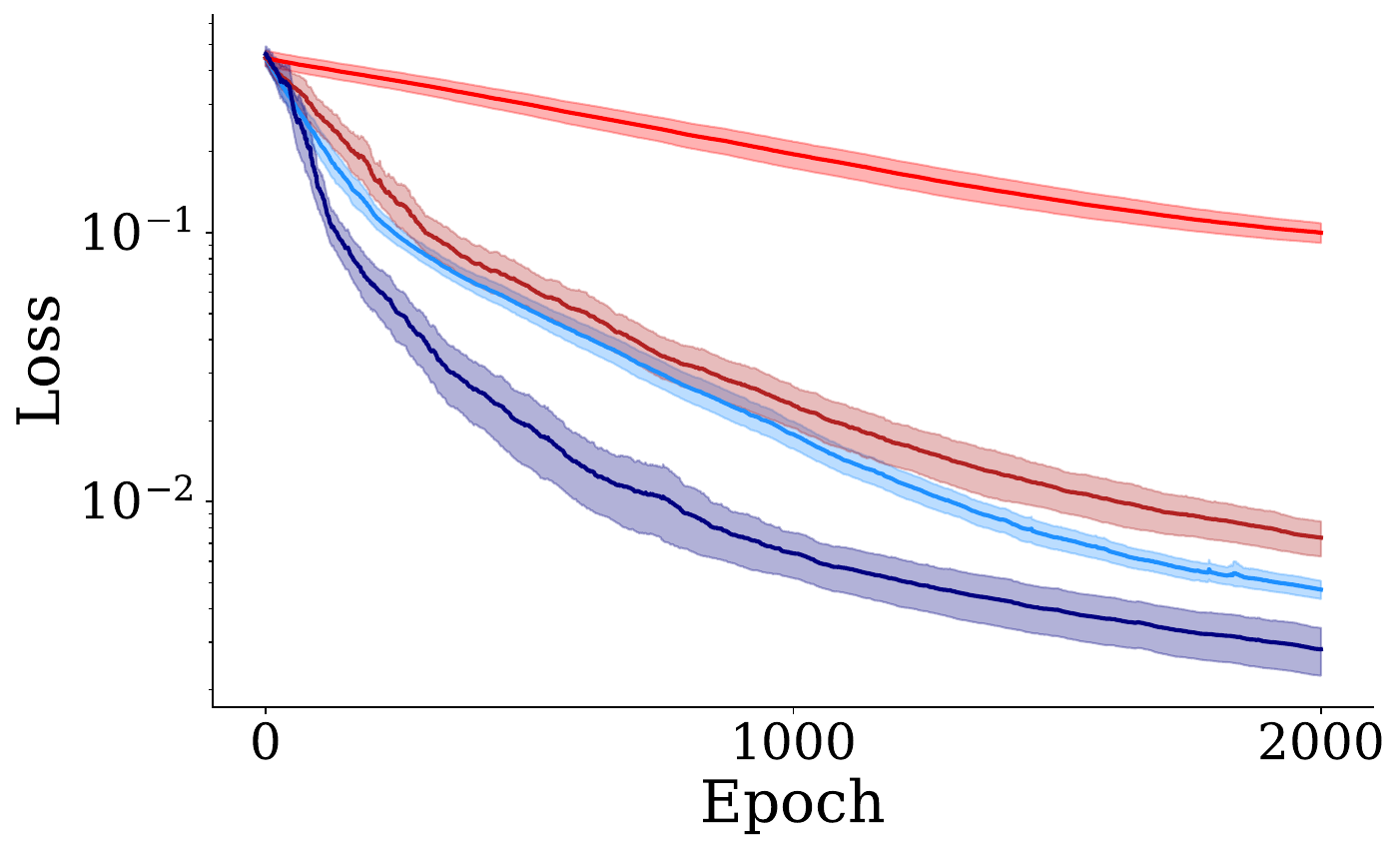} 
        \label{fig:rossler_main1}
    \end{subfigure}
    \begin{subfigure}[t]{0.32\textwidth}
        \centering
        \includegraphics[width=1.\linewidth]{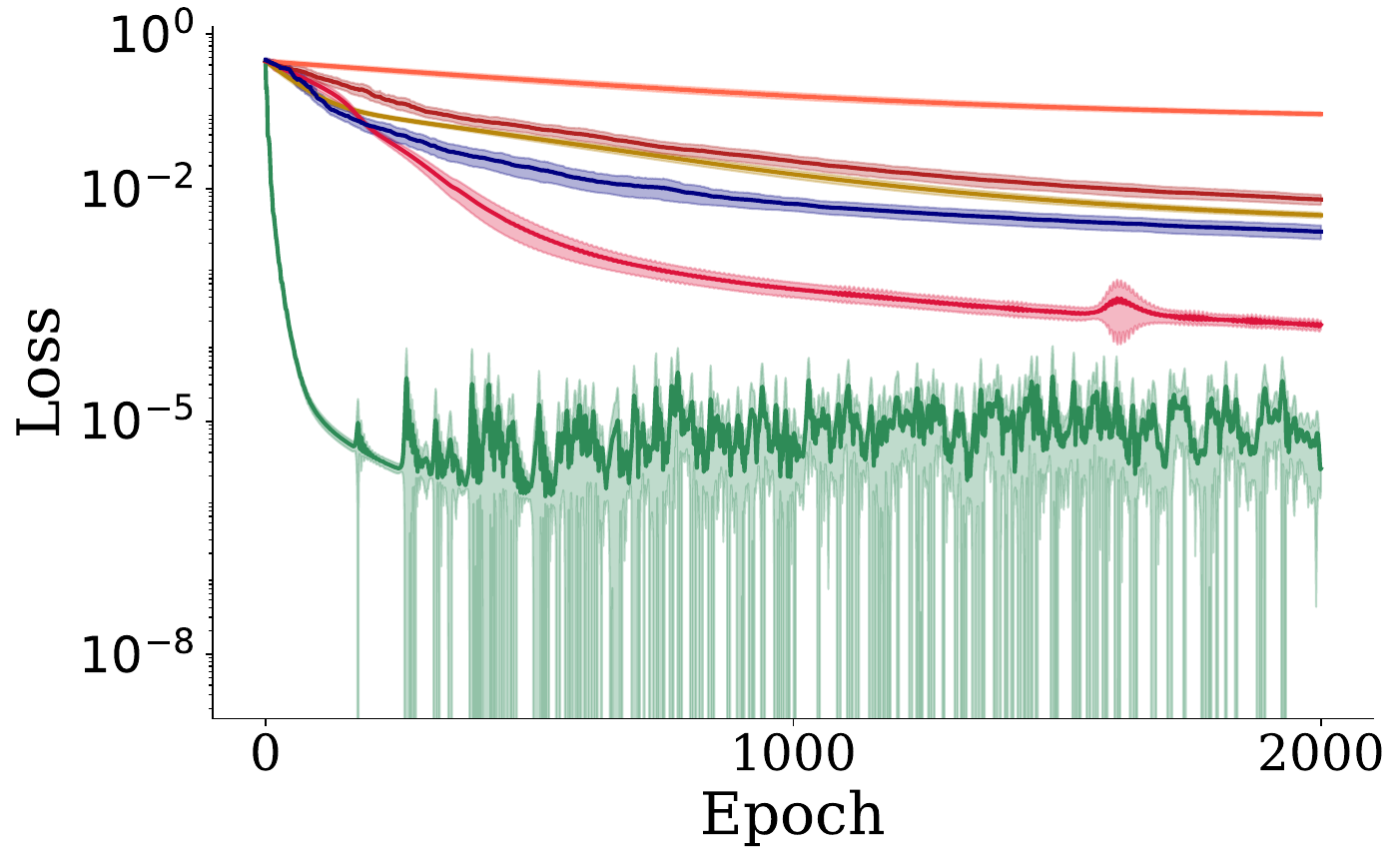} 
        \label{fig:rossler_main2}
    \end{subfigure}
    \hfill{}\hspace{0.1\textwidth}
    \scalebox{0.75}{
    \begin{subfigure}[t]{0.32\textwidth}
        \centering
        \vspace{-12em}
        \hspace{-12em}
        \begin{tabular}{c|c} 
        \textbf{Rossler} & Average final loss; (CI) \\ 
        \hline
        SGD & 0.0046; (0.0042, 0.0050) \\ 
        Adam & $1e^{-6}$; ($1e^{-6}$, $1e^{-6}$) \\ 
        DNI-SGD & 0.0926; (0.0869, 0.0982) \\
        DNI-Adam & 0.0002; (0.0002, 0.0002) \\ 
        WP & 0.0999; (0.0915, 0.1083) \\ 
        SWP & 0.0047; (0.0043, 0.0051) \\ 
        Dopamine1 & 0.0073; (0.0063, 0.0084) \\ 
        Dopamine2 & 0.0028; (0.0023, 0.0034) \\ 
        \end{tabular}
    \end{subfigure}}
    \vspace{-10pt}
    \caption{Training performance of RNN models under various optimization methods on Lorenz attractor system (\textit{top}) and Rössler attractor systems (\textit{bottom}) where the Loss in $y$-axis plotted in logarithmic scale. The columns of the table report the average final loss and (in brackets) its 95$\%$ Confidence Interval (CI) calculated from 20 separate training runs with different random seeds. See \autoref{tab:lorenz_hyperparam} and \autoref{tab:rossler_hyperparam} for more details on hyperparameters.}  
    \label{fig:ts_forecasting_performance}
\end{figure*}

\vspace{-0.5em}
\subsection{Recurrent Neural Networks: Chaotic time-series forecasting task}

Predicting time series from chaotic dynamical systems has served as a benchmark in modeling extreme events and has been used to probe the behavior of various optimization methods and architectures in Deep Learning \cite{brunton2016discovering,kaiser2018sparse, han2019review}. Hence, for the time series forecasting task, we used trajectories drawn from two classical dynamical system models, the Rössler \cite{rossler1983chaotic} and the Lorenz strange attractor \cite{lorenz1963deterministic} with single and double spirals respectively (see \autoref{apx:rossler_eqn} for more details on the Rössler and Lorenz attractors equations used to generate the target chaotic trajectories). We used RNNs with 512 units with ReLU nonlinearity and a linear readout layer. The network is trained for 2000 iterations with a batch size of 5000 and a look-back window size of 32. The objective is to minimize the Mean Squared Error (MSE) between the predictions and the target trajectories using the optimizers under study. 

\begin{figure}[!t]
    \begin{minipage}[t]{\columnwidth}
        \centering
        \includegraphics[width=0.8\linewidth]{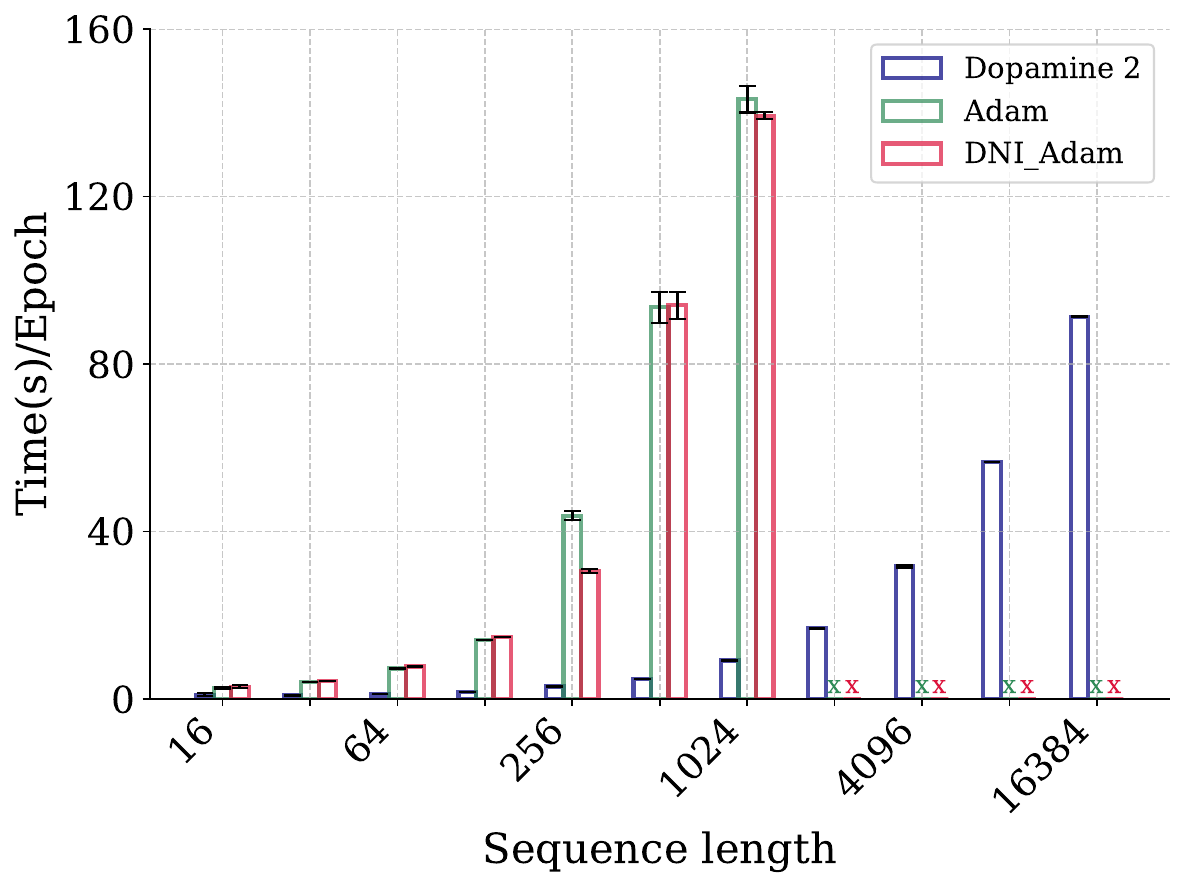}
        \label{fig:time_forward_backward}
    \end{minipage}
    
    \begin{minipage}[t]{\columnwidth}
        \centering
        \includegraphics[width=0.8\linewidth]{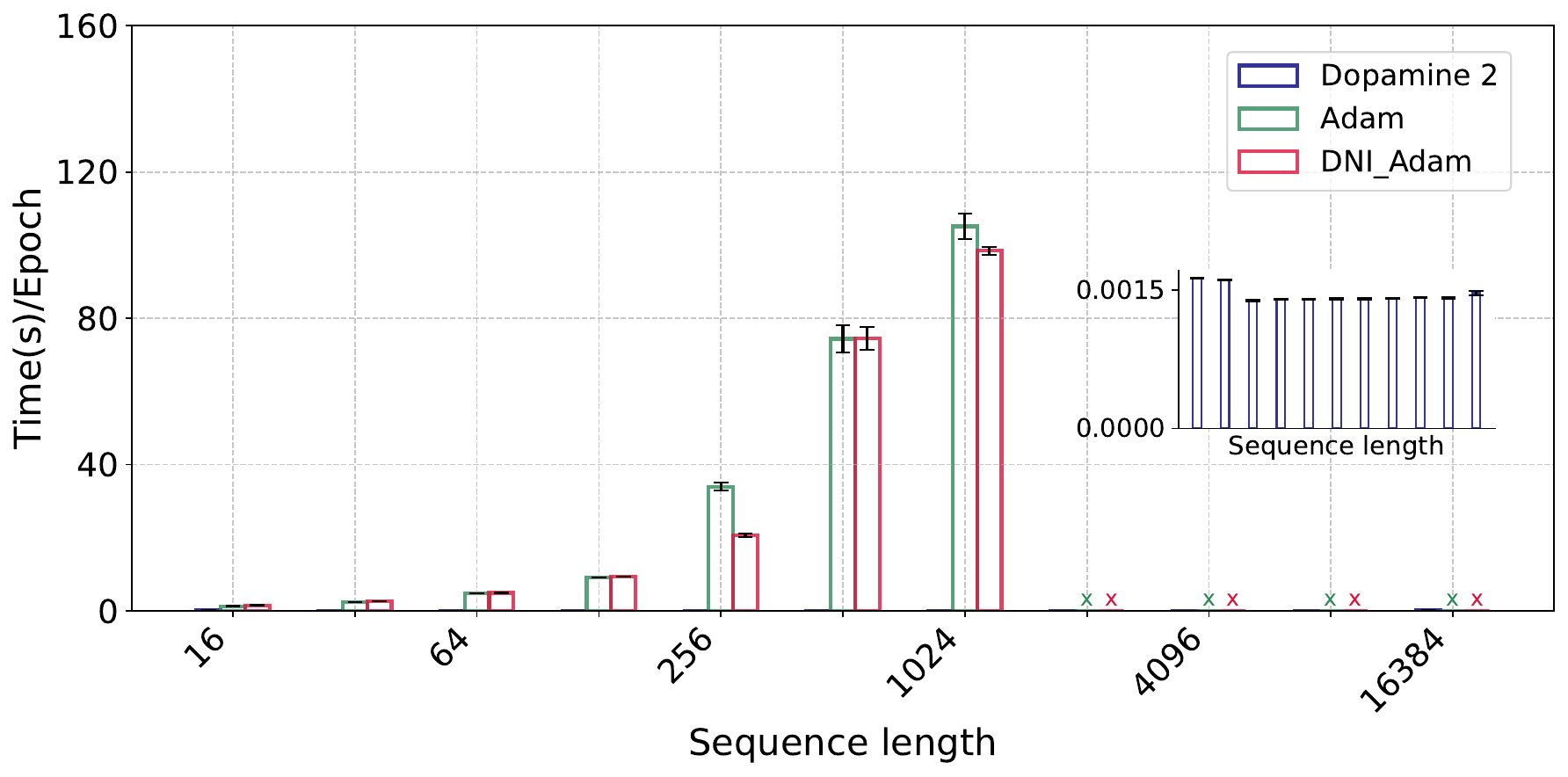}
        \label{fig:time_backward}
    \end{minipage}
    \vspace{0.1em}

    \begin{minipage}[t]{\columnwidth}
        \centering
        \scriptsize
        \begin{tabularx}{0.55\linewidth}{@{}lrr@{}}
            \toprule
            \multicolumn{3}{c}{\textbf{Backward Pass}} \\
            \midrule
            Optimizer & \multicolumn{1}{c}{Mean ± SEM} & \multicolumn{1}{c}{Median} \\
            \midrule
            Adam & 105.15 ± 3.40 & 95.73 \\
            Dopamine 2 & 0.0014 ± 5e-6 & 0.0014 \\
            DNI-Adam & 98.48 ± 1.11 & 96.60 \\
            \addlinespace[0.5em]
            \multicolumn{3}{c}{\textbf{Forward + Backward}} \\
            \midrule
            Optimizer & \multicolumn{1}{c}{Mean ± SEM} & \multicolumn{1}{c}{Median} \\
            \midrule
            Adam & 143.32 ± 3.21 & 135.26 \\
            Dopamine 2 & 9.08 ± 0.09 & 8.93 \\
            DNI-Adam & 139.36 ± 0.79 & 138.45 \\
            \bottomrule
        \end{tabularx}
    \end{minipage}
    \vspace{0.3cm}
    \caption{Time and computational complexity of gradient-based optimizers and Dopamine. (\textit{Top}) The time taken for both forward and backward passes, (\textit{Middle}) time for backward pass alone, and (\textit{Bottom}) corresponding timing data table. Dopamine shows constant negligible time across sequence lengths (mean, SEM, and median in table). Crosses indicate optimizer failures at specific sequence lengths. Table shows timing for e.g., sequence length 1024, with full results in \autoref{tab:time_epoch_backward} and \autoref{tab:time_epoch_all}. See \autoref{appx:hpc} for hardware details.}
    \vspace{-0.5cm}
    \label{fig:time_computation}
\end{figure}

Our results show that RNN models trained using Dopamine exhibit accelerated convergence compared to BPTT-SGD and standard WP and fit the underlying dynamical trajectory approximately (see \autoref{fig:lorenz_rossler_predictions} for dopamine 2, \autoref{fig:adaptive_lr_lorenz_rossler} for dopamine 1, and  \autoref{fig:ts_forecasting_performance} right column for quantitative assessments). Moreover, while, BPTT-Adam variants (including DNI) show slightly faster convergence and better minima (see also \autoref{fig:ts_forecasting_performance}), \textit{Dopamine} surpasses them in both computation and memory efficiency (\autoref{fig:time_computation}). To demonstrate this, we took 50k time points from the Rössler attractor trajectory and trained the RNN model independently with varying sequence lengths. We recorded the clock time (in seconds) taken by the optimizers during the backward path and also for both the forward and backward paths. We found that BPTT optimizers, due to their dependence on sequence length, showed exponential growth in the time per backward iteration, whereas Dopamine maintained constant duration regardless of the sequence length. 

Moreover, during the forward pass, it is notable that Dopamine can handle sequence lengths greater than BPTT optimizers due to its lower computational complexity (\autoref{fig:time_computation}). To assess that, we used a fixed memory resource (\autoref{appx:hpc}) for all training algorithms. This advantage stems from Dopamine's derivative-free optimization. Given $N$ number of neurons in the RNN and sequence length of T, Dopamine has reduced memory complexity to $O(N)$ compared to gradient-based methods, which have a complexity of $O(NT)$. Moreover, Dopamine's parameter update operations are parallelizable and independent of time $T$, resulting in a total computational complexity of $O(2NT+\boldsymbol{N})$ given a sequence of length $T$ for $N$ units and it requires 2 forward computations to acquire RPE. In contrast, the computational complexity of BPTT is $O(NT+TN^2)$ due to the process of unrolling the network over time(T) during the backward pass. 
These findings suggest that the Dopamine Optimizer provides a computationally efficient algorithm for training RNNs in general.

\section{Discussions}

The key leverage of our Dopamine optimizer is using local learning rules and adaptive learning rate, where weight modifications depend only on the direction and magnitude of the global reward signal. This not only brings resource efficiency with a comparable performance, but also further biological plausibility for training neural networks. Such a synaptic weight update can also be used to train spiking neural networks as it was shown that the global error signal has a distinct impact on the spiking behavior \cite{richards2019dendritic, sacramento2018dendritic}.

Furthermore, in our training process, state transitions and rewards follow the Markov property (it is memoryless), since the future outcomes depend only on the current state, not on the history. Thus, further scaling the learning rate based on the magnitude and direction of the regret allows faster convergence at tolerable approximation error compared to other perturbation-based methods, such as SOTA  algorithms. 

Overall, under finite difference stochastic approximation using weight perturbation settings, we introduced adaptive learning rate as a function of the moving average of the RPE or regret. Simulation results showed accelerated learning, better generalization (by avoiding saddle points), and reduced computational demand and memory usage compared to gradient-based optimizers such as SGD and Adam, as well as more biological plausibility.

\section{Limitations}
\label{sec:limit}

While our optimizer, Dopamine, has various advantages, there are several directions that can be further improved.  
First, Dopamine's performance relies on its adaptive learning rate, which is determined by a set of hyperparameters (introduced in Equations \ref{eqn:s_stochastic}-\ref{eqn:eta_exponential_decay}).
Through our proposed heuristics we can find these parameters to reach high performance, however, more theoretically grounded approaches are needed. 
Second, even after a long training (e.g., 2000 epochs), a small residual error persists. 
This does not only apply to Dopamine optimizers but also to others (see \autoref{fig:ts_forecasting_performance}), however, in the case of Dopamine, 
we conjecture that the bias is attributed to the variance of the noise ($\xi$) we used to perturb the model parameters. This residual error can be potentially reduced, by using an adaptive noise.

\bibliography{main}
\bibliographystyle{icml2021}


\newpage

\onecolumn
\appendix

\setcounter{figure}{0}  
\renewcommand{\thefigure}{S\arabic{figure}}
\renewcommand{\thetable}{S\arabic{table}}
\renewcommand{\theHtable}{Supplement.\thetable}
\renewcommand{\theHfigure}{Supplement.\thefigure}

\section{Supplemental material}

\subsection{Spectral Radius}\label{appx:spectral_radius}

Let us assume that the recurrent weights $\theta^{rec}$ of the RNN are diagonalizable and its eigen decomposition $$\theta^{rec}=Q \Lambda Q^{-1}$$ exists, where Q contains the eigenvectors and the diagonal matrix $\Lambda$ holds the spectrum of eigenvalues. Then the spectral radius of $\theta^{rec}$ is defined as the supremum of the absolute eigenvalues in the spectrum,
$$\rho(\theta^{rec}) = \sup_{\lambda \in \Lambda(\theta^{rec})}|\lambda|$$

In order to modify the spectral radius $\rho(\theta^{rec})$ of the recurrent weights to a desired value $\hat{\lambda}$, we multiply $\theta^{rec}$ with the ratio $\frac{{\hat{\lambda}}}{\rho(\theta^{rec})}$. For the sake of clarity, we keep the notation $\lambda$ as the desired spectral radius. 

\subsection{XOR Classification Task}

\begin{table}[h]
	\centering
	\begin{tabular}{c|c|c|c|c|c|c}
		
		& $\eta_0$ & $s_0$ & $\beta_s$ & $\beta_\eta$  & $\sigma^{2}$ \\
		\hline 
		WP & $1e^{-3}$ & $-$ & $-$ & $-$ &   $1e^{-1}$ \\
		\hline
		Dopamine 1 & $1e^{-2}$ & $1e^{-2}$ & $0.001$ &    $1e^{-4}$  & $1e^{-1}$ \\
		\hline
		Dopamine 2 & $1e^{-1}$ & $1e^{-1}$ & $0.9998$ & $0.999$  & $1e^{-1}$ \\
		\hline
		 
	\end{tabular}
    \vspace{0.3cm}
	\caption{Hyperparameter values of optimizers used in XOR classification task:$\eta_0$ denotes the initial learning rate, $s_0$ denotes the initial value of the auxiliary variable, the moving average of the regret score $\mathcal{R}$ and the constant coefficients $\beta_s$ and $\beta_{\eta}$ of the \textit{Dopamine} optimizers. The final performance of the gradient based optimizers are insensitive to the initial learning rate, hence we fixed it to $1e^{-3}$.}
	\label{tab:xor_hyperparam}
\end{table}
\vspace{1em}

\begin{table}[h]
	\centering
	\begin{tabular}{c|c|c|c|c|c|c}
		& $\eta_0$ & $s_0$ & $\beta_s$ & $\beta_\eta$  & $\sigma^{2}$ \\
        \hline
		Dopamine 1a & $1e^{-2}$ & $1e^{-2}$ & $0.001$ & $0.0001$  & $1e^{-1}$ \\
        Dopamine 1b & $1e^{-2}$ & $1e^{-2}$ & $0.0001$ & $0.001$  & $1e^{-1}$ \\
        Dopamine 1c & $1e^{-2}$ & $1e^{-2}$ & $0.0001$ & $0.005$  & $1e^{-1}$ \\
        Dopamine 2a & $1e^{-1}$ & $1e^{-1}$ & $0.9998$ & $0.0001$  & $1e^{-1}$ \\
        Dopamine 2b & $1e^{-1}$ & $1e^{-1}$ & $0.9998$ & $0.001$  & $1e^{-1}$ \\
        Dopamine 2c & $1e^{-1}$ & $1e^{-1}$ & $0.9998$ & $0.999$  & $1e^{-1}$ \\
		\hline
		 
	\end{tabular}
        \vspace{0.3cm}
	\caption{Hyperparameter values of optimizers used for ablation studies:$\eta_0$ denotes the initial learning rate, $s_0$ denotes the initial value of the auxiliary variable, the moving average of the regret score $\mathcal{R}$ and the constant coefficients $\beta_s$ and $\beta_{\eta}$ of the \textit{Dopamine} optimizers.}
	\label{tab:xor_hyperparam_ablation}
\end{table}

\begin{table}[h]
	\centering
	\begin{tabular}{cccccccc}
    \hline
    RNN Layers & Hidden Units & Epochs & Batch Size & Sequence Length & Data Size & Non-linearity & $\sigma^{2}$  \\
    \hline
    1 & 512 & 5000 & 2000 & 32 & 5000 & ReLU &  $1e^{-5}$ \\
    \hline
    \end{tabular}
    \vspace{0.3cm}
	\caption{Hyperparameter values of Dopamine 2 used for ablation studies:$\eta_0$ denotes the initial learning rate, $s_0$ denotes the initial value of the auxiliary variable, the moving average of the regret score $\mathcal{R}$ and the constant coefficients $\beta_s$ and $\beta_{\eta}$ of the \textit{Dopamine 2} optimizer.}
	\label{tab:rossler_hyperparam_ablation}
\end{table}

\subsection{Hardware specifics}

\subsubsection{Desktop}\label{appx:desktop}
\textit{GPU}: NVIDIA RTX 4060 - 16GB \\
\textit{Processor}: 13th Gen Intel(R) Core(TM) i7-13650HX,
2600 Mhz, 14 Core(s), 20 Logical Processor(s)\\
\textit{Physical Memory (RAM)}: 16.0 GB
Experiments: XOR classification task

\subsubsection{High Peformance Computing(HPC) cluster }\label{appx:hpc}

\textit{Processors}: 2 x Intel Xeon Platinum 8470 (52 cores) @ 2.00 GHz \\
\textit{Physical Memory (RAM)}: 512 GB RAM (8 x 32 GB DDR5-4800 MT/s per socket)
Experiments: Chaotic time series prediction task

\subsection{Step-ahead prediction task}
\subsubsection*{Rossler Attractor Equation}\label{apx:rossler_eqn}
Evolution of state variables $x,y$ and $z$ of the Rossler attractor can be described by the following differential equations; 
\begin{equation}
    \dot{x}_t= -(y_t+z_t)
\end{equation}
\begin{equation}
    \dot{y}_t = x_t+ay_t
\end{equation}
\begin{equation}
    \dot{z}_t = b+x_tz_t-cz_t
\end{equation}
where $a$,$b$ and $c$ are the chaotic real parameters with fixed values $0.2$,$0.2$ and $5.7$. To generate the trajectories used in this study, the state variables are initialized at $x_0,y_0$ and $z_0$ are initialized at $1.0$, $0.0$ and $0.0$ respectively.
\subsubsection*{Lorenz Attractor Equation}\label{apx:lorenz_eqn}
Lorenz chaotic attractor can be described by the following set of differential equations;
\begin{equation}
    \dot{x}_t= \sigma(x_t-y_t)
\end{equation}
\begin{equation}
    \dot{y}_t = \rho x_t-x_tz_t
\end{equation}
\begin{equation}
    \dot{z}_t = \beta_ty_t-bz_t
\end{equation}
To simulate the trajectory used for step-ahead prediction task, we set the chaotic parameters as $\sigma=10$,$\rho=28$, $\beta = 8/3$ and $dt=0.01$. The state variables $x_0,y_0$ and $z_0$ are initialized at $1.0$, $0.0$ and $0.0$ respectively. Trajectories used in this study where sampled by solving the above state equations iteratively for $5000$ time steps. 
\subsection{Hyperparameters for step-ahead prediction tasks}
\begin{table}[h]
	\centering
	\begin{tabular}{c|c|c|c|c|c|c}
		
		& $\eta_0$ & $s_0$ & $\beta_s$ & $\beta_\eta$ & $\lambda$ & $\sigma^{2}$ \\
		\hline 
		WP & $1e^{-3^{*}}$ & $-$ & $-$ & $-$ & $-$ &  $1e^{-4}$ \\
		\hline
		Spectral WP & $1e^{-2}$ & $-$ & $-$ & $-$ & $1.0$ & $1e^{-4}$ \\
		\hline
		Dopamine 1 & $1e^{-2}$ & $1e^{-4}$ & $0.9998$ & $1e^{-4}$ & $1.0$ & $1e^{-4}$ \\
		\hline
		Dopamine 2 & $1e^{-1}$ & $3e^{-4}$ & $0.9998$ & $0.999$ & $1.0$ & $1e^{-4}$ \\
		\hline
		 
	\end{tabular}
    \vspace{0.3cm}
	\caption{Hyperparameter values of reward-based optimizers used in forecasting Lorenz attractor trajectory with RNN:$\eta_0$ denotes the initial learning rate; $s_0$ denotes the initial value of the auxiliary variable, the moving average of the regret score $\mathcal{R}$ and the constant coefficients $\beta_s$ and $\beta_{\eta}$ of the \textit{Dopamine} optimizers, $\lambda$ - the spectral radius of the recurrent weights, fixed to $1.0$. The scale of the perturbation $\sigma$ is fixed to $0.0001$. ${}^{*}$Learning rate is adjusted (reduced) to avoid gradient explode.} 
	\label{tab:lorenz_hyperparam}
\end{table}
\begin{table}[h]
	\centering
	\begin{tabular}{c|c|c|c|c|c|c}
		
		& $\eta_0$ & $s_0$ & $\beta_s$ & $\beta_\eta$ & $\lambda$ & $\sigma^{2}$ \\
		\hline 
		WP & $1e^{-4^{*}}$ & $-$ & $-$ & $-$ & $-$ &  $1e^{-5}$ \\
		\hline
		Spectral WP & $1e^{-3^{*}}$ & $-$ & $-$ & $-$ & $1.0$ & $1e^{-5}$ \\
		\hline
		Dopamine 1 & $1e^{-2}$ & $1e^{-4}$ & $0.9998$ & $1e^{-4}$ & $1.0$ & $1e^{-5}$ \\
		\hline
		Dopamine 2 & $1e^{-2}$ & $1e^{-4}$ & $0.99998$ & $1e^{-5}$ & $1.0$ & $1e^{-5}$ \\
		\hline	
	\end{tabular}
    \vspace{0.3cm}
	\caption{Hyperparameter values of optimizers used in forecasting Rössler attractor trajectory using RNN:$\eta_0$ denotes the initial learning rate; $s_0$ denotes the initial value of the auxiliary variable, the moving average of the regret score $\mathcal{R}$ and the constant coefficients $\beta_s$ and $\beta_{\eta}$ of the \textit{Dopamine} optimizers and $\lambda$ - the spectral radius of the recurrent weights is fixed to $1.0$. The scale of the perturbation $\sigma$ is fixed to $0.00001$. ${}^{*}$Learning rate is adjusted (reduced) to avoid gradient explode.}
	\label{tab:rossler_hyperparam}
\end{table}

\vspace{1em}

\begin{figure*}[!b]

    \centering
    \begin{subfigure}[b]{0.24\textwidth}
    	\captionsetup[subfigure]{position=top}
        \centering
        \caption*{BP-Adam}
        \includegraphics[width=\textwidth, height=0.7\textwidth]{imgs/ls/loss_surface_adam_affine_train.pdf}
    \end{subfigure}
    \hfill
    \begin{subfigure}[b]{0.24\textwidth}
        \centering
        \caption*{DNI(linear)-Adam}
        \includegraphics[width=\textwidth, height=0.7\textwidth]{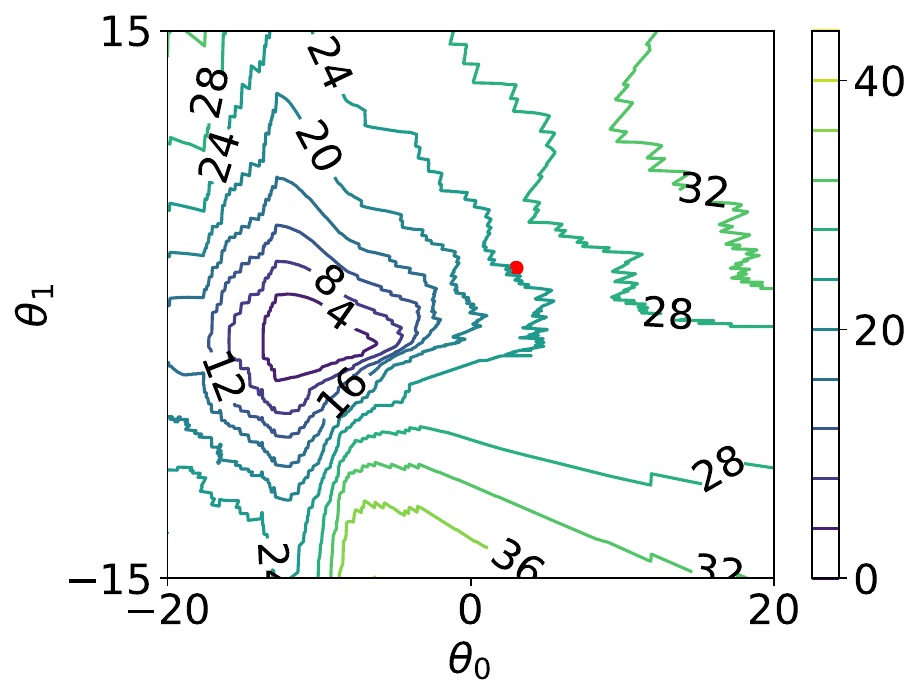}
    \end{subfigure}
    \hfill
    \begin{subfigure}[b]{0.24\textwidth}
        \centering
        \caption*{Dopamine-$1$}
        \includegraphics[width=\textwidth, height=0.7\textwidth]{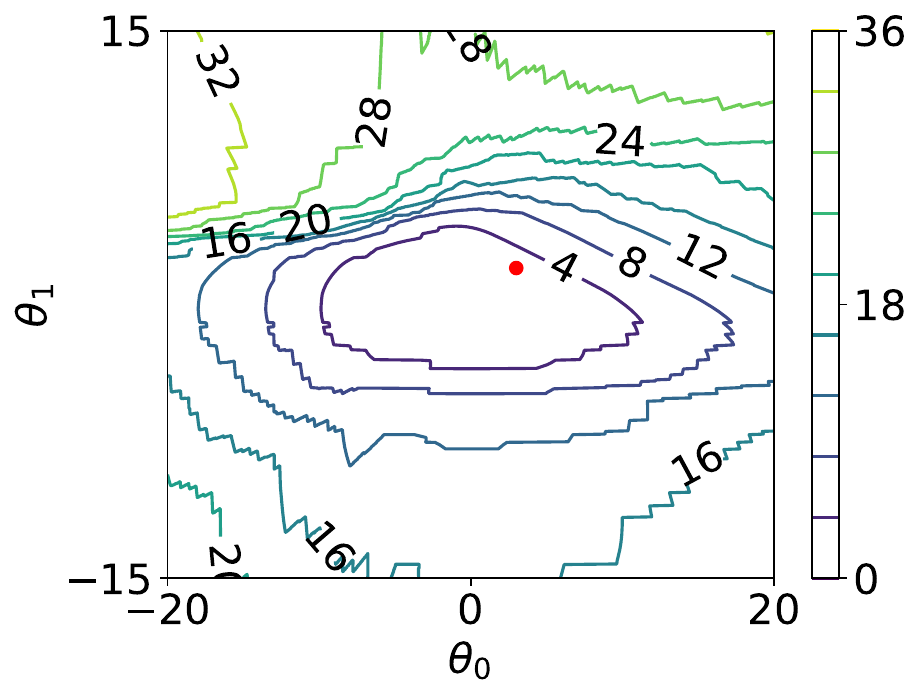}
    \end{subfigure}
	\hfill
	\begin{subfigure}[b]{0.24\textwidth}
		\centering
    		\caption*{Dopamine-$2$}
		
		\includegraphics[width=\textwidth, height=0.7\textwidth]{imgs/ls/loss_surface_dopamine2_affine_train.pdf}%

	\end{subfigure}
	\hfill
    \begin{subfigure}[b]{0.24\textwidth}
        \centering   
        \includegraphics[width=\textwidth, height=0.7\textwidth]{imgs/ls/loss_surface_adam_affine_test.pdf}
    \end{subfigure}
    \hfill
    \begin{subfigure}[b]{0.24\textwidth}
        \centering
        \includegraphics[width=\textwidth, height=0.7\textwidth]{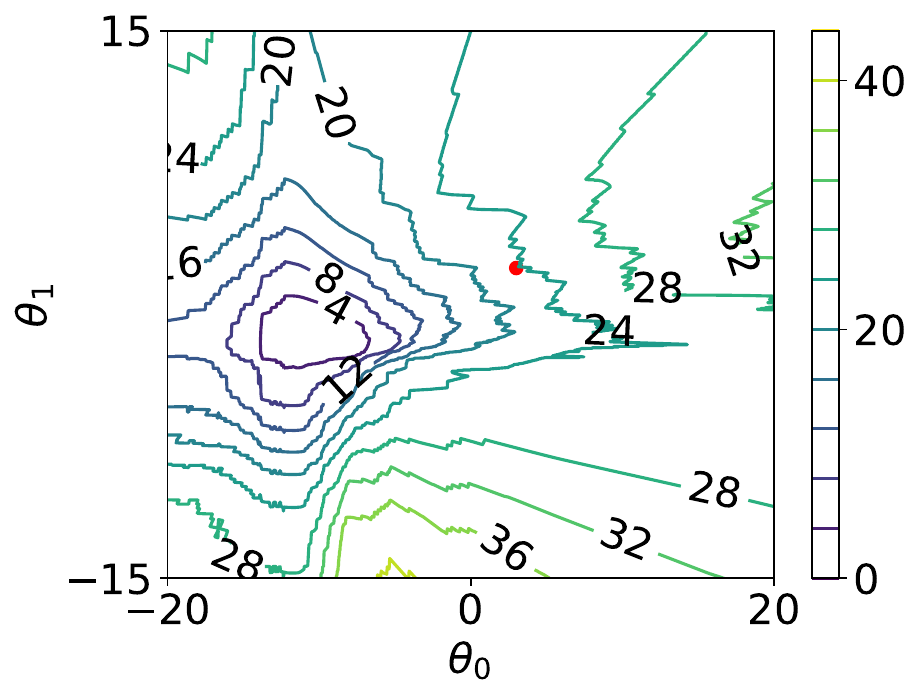}
    \end{subfigure}
    \hfill
    \begin{subfigure}[b]{0.24\textwidth}
        \centering
        \includegraphics[width=\textwidth, height=0.7\textwidth]{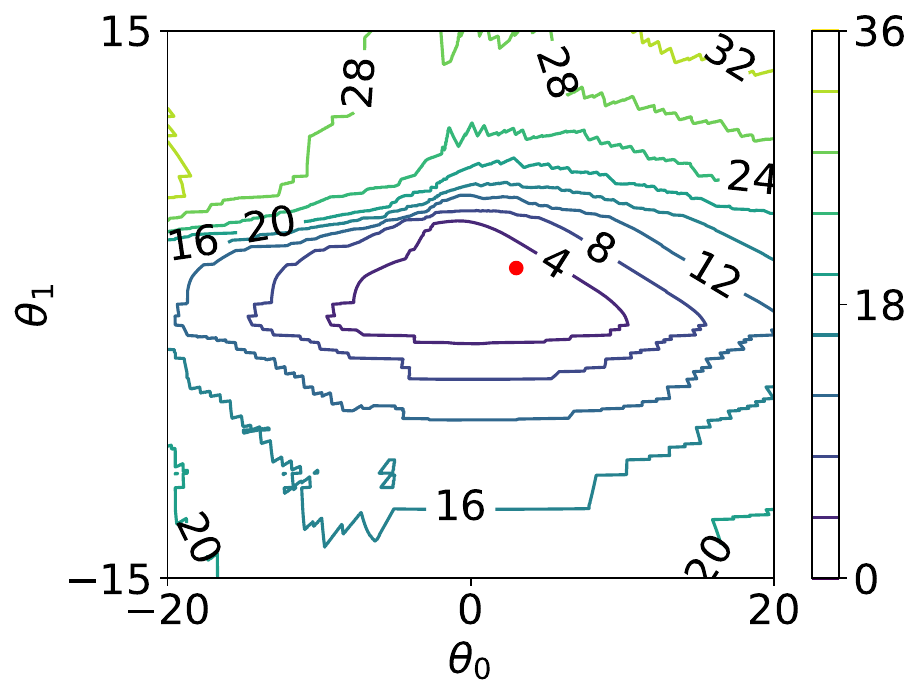}
    \end{subfigure}
	\hfill
	\begin{subfigure}[b]{0.24\textwidth}
		\centering
		
		\includegraphics[width=\textwidth, height=0.7\textwidth]{imgs/ls/loss_surface_dopamine2_affine_test.pdf}
		
	\end{subfigure}
	\hfill
    \caption{XOR classification task: Loss Landscape computed using random direction method shows Dopamine trained models have smaller generalization gap. See \autoref{tab:xor_hyperparam} for hyperparameter details.}
    \label{fig:loss_landscape}
\end{figure*}

\begin{figure*}[!t]
    \centering
    
    \includegraphics[width=0.45\textwidth]{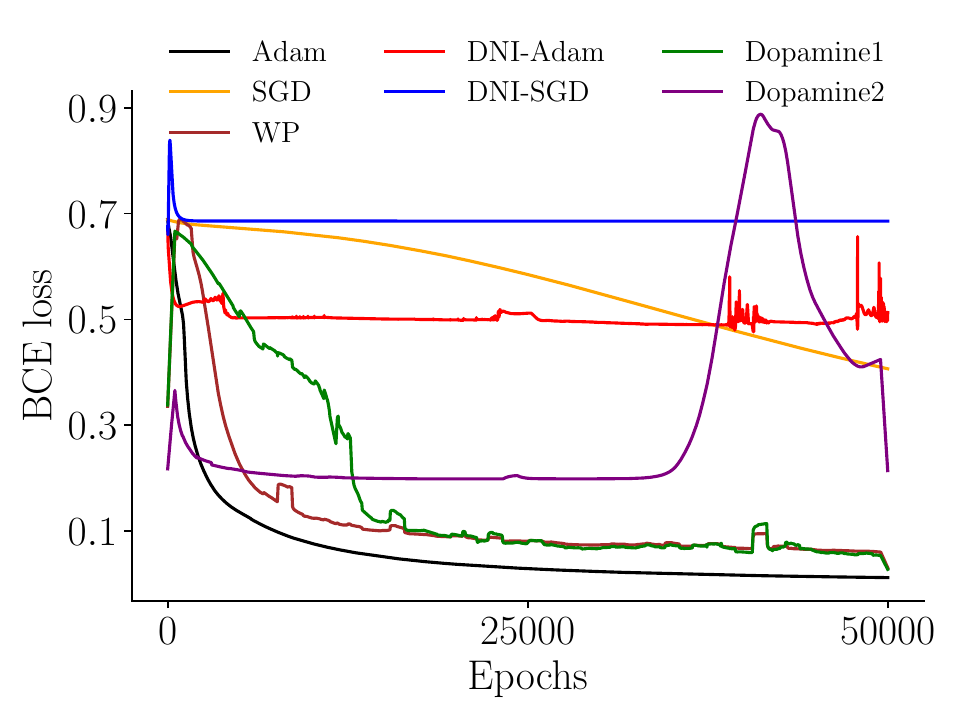}%
    \label{fig:xor_losses}%
    
    \caption{Training performance of various optimizers on XOR classification problem.}
    \label{fig:xor_loss}
\end{figure*}

\begin{table}[h]
	\centering
	\begin{tabular}{c|c|c|c|c|}
		
		& SGD & Adam & DNI-SGD & DNI-Adam \\
		\hline 
		Lorenz & $1e^{-3}$ & $1e^{-3}$ & $1e^{-4^{*}}$ & $1e^{-5^{*}}$ \\
		\hline
		Rössler & $1e^{-3}$ & $1e^{-3}$ & $1e^{-4}$ & $1e^{-4}$ \\
		\hline
	\end{tabular}
    \vspace{0.3cm}
	\caption{Initial $\eta_0$ of BPTT optimizers for Lorenz and Rössler chaotic trajectory forecasting task: Other network and training hyperparameters such as RNN size(512), Sequence length(32), batch size(5k) and epochs(2k) stay constant across the datasets and optimizers.${}^{*}$Learning rate is adjusted(reduced) to avoid gradient explode.} 
	\label{tab:lorenz_rossler_sgd_hyperparam}
\end{table}
 
\begin{figure*}[!b]
	\centering
	\begin{subfigure}[b]{0.24\textwidth}
		\captionsetup[subfigure]{position=top}
		\centering
		\caption*{BP-Adam}
		\includegraphics[width=\textwidth, height=0.7\textwidth]{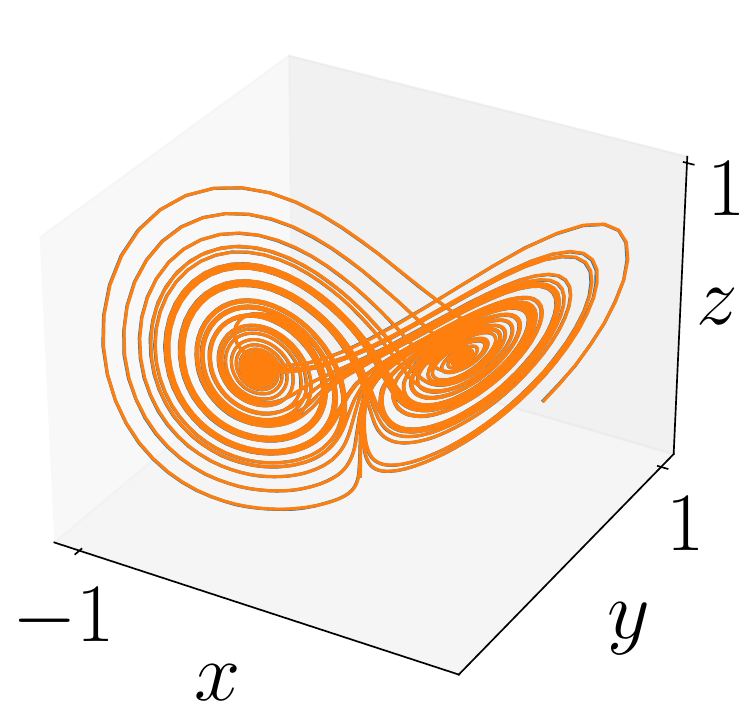}
	\end{subfigure}
	\hfill
	\begin{subfigure}[b]{0.24\textwidth}
		\centering
		\caption*{DNI-SGD}
		\includegraphics[width=\textwidth, height=0.7\textwidth]{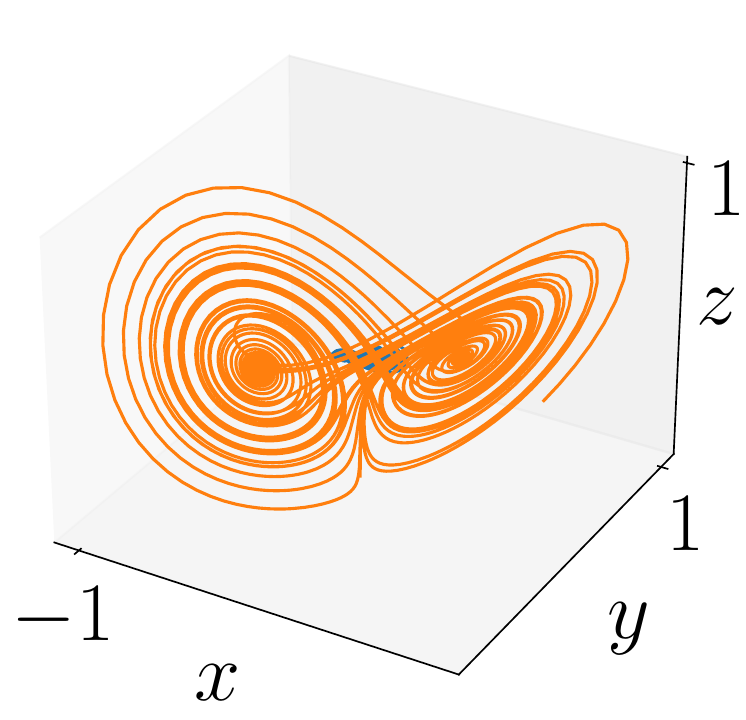}
	\end{subfigure}
	\hfill
	\begin{subfigure}[b]{0.24\textwidth}
		\centering
		\caption*{DNI-Adam}
		\includegraphics[width=\textwidth, height=0.7\textwidth]{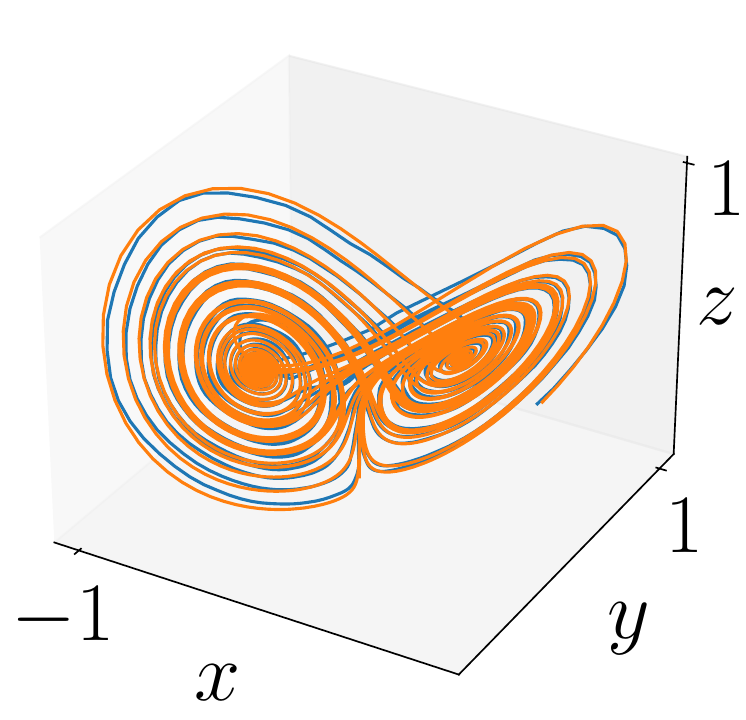}
	\end{subfigure}
	\hfill
	\begin{subfigure}[b]{0.24\textwidth}
		\centering
		\caption*{Dopamine 1}
		\includegraphics[width=\textwidth, height=0.7\textwidth]{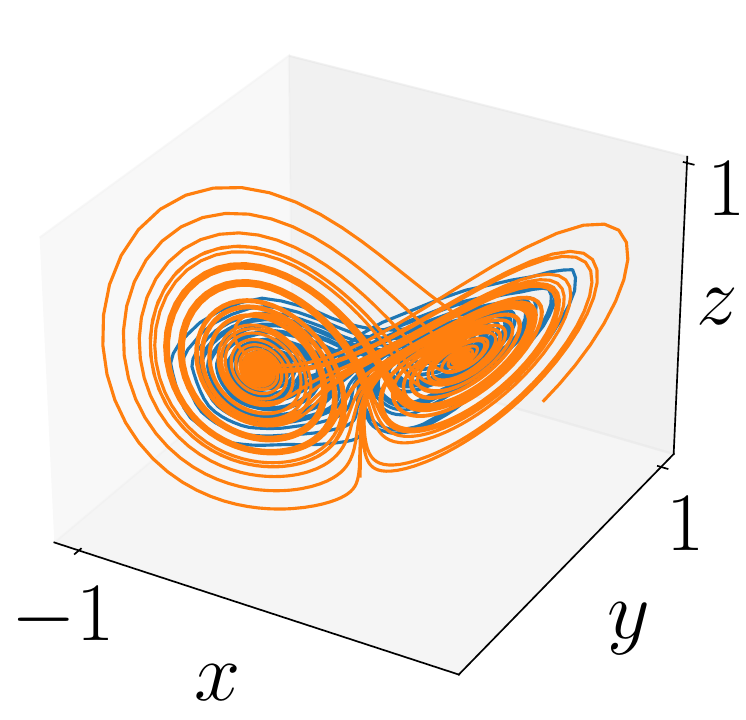}%

	\end{subfigure}
	\hfill
	\begin{subfigure}[b]{0.24\textwidth}
		\centering   
		\includegraphics[width=\textwidth, height=0.7\textwidth]{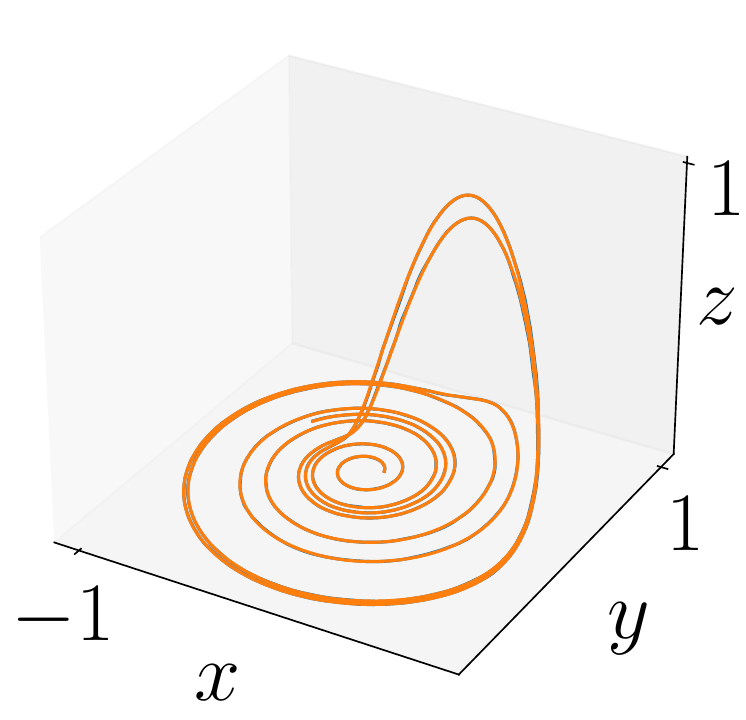}
	\end{subfigure}
	\hfill
	\begin{subfigure}[b]{0.24\textwidth}
		\centering
		\includegraphics[width=\textwidth, height=0.7\textwidth]{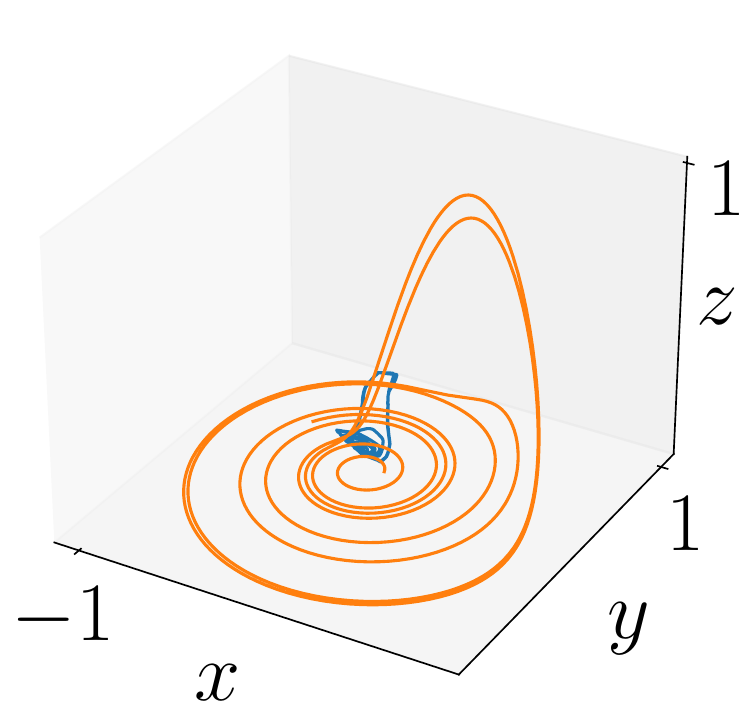}
	\end{subfigure}
	\hfill
	\begin{subfigure}[b]{0.24\textwidth}
		\centering
		\includegraphics[width=\textwidth, height=0.7\textwidth]{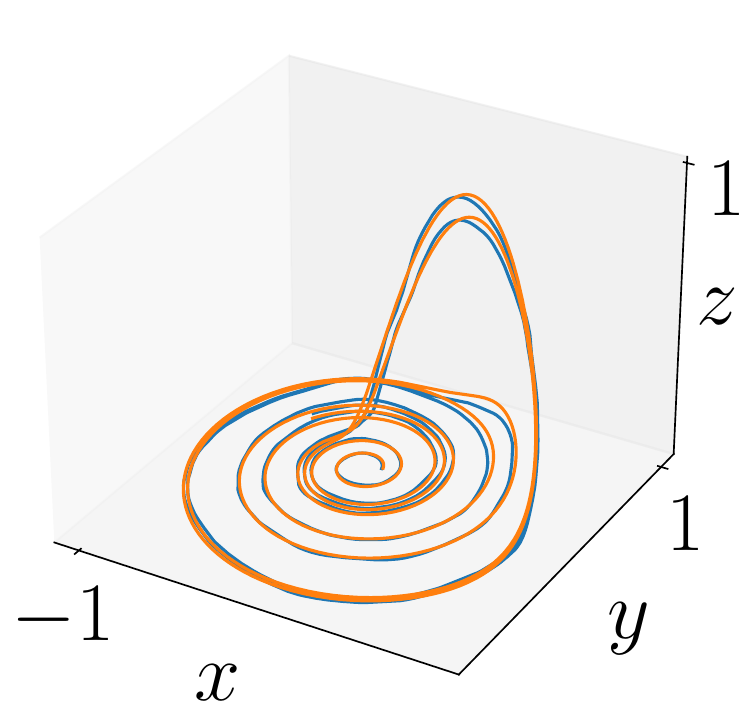}
	\end{subfigure}
	\hfill
	\begin{subfigure}[b]{0.24\textwidth}
		\centering
		\includegraphics[width=\textwidth, height=0.7\textwidth]{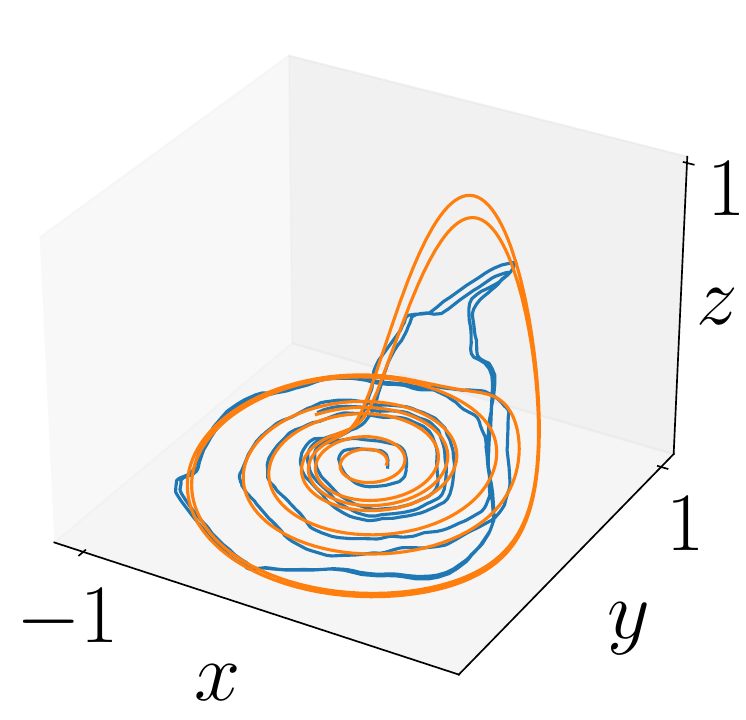}
		
	\end{subfigure}
	\hfill
    \vspace{0.3cm}
	\caption{Predictive performance of RNN models trained using various optimizers. See \autoref{tab:rossler_hyperparam} and \autoref{tab:lorenz_rossler_sgd_hyperparam}  for hyperparameter details, DNI SGD LR = 1e-4, Adam = 1e-4, and LR of Spectral WP for lorenz =1e-2 and rossler = 1e-3.}
 \label{fig:adaptive_lr_lorenz_rossler}
\end{figure*}

\begin{table}[h]
	\centering
	\begin{tabular}{c|c|c|c|c|c|c|}
		
		Look-back window & \multicolumn{2}{c|}{Adam} & \multicolumn{2}{c|}{DNI-Adam} & \multicolumn{2}{c|}{Dopamine 2} \\
        \cline{2-7}
        & Mean ± SEM & Median & Mean ± SEM & Median & Mean ± SEM & Median \\
		\hline 

		16 & $1.3088\pm0.0930$ & $1.1767$ & $1.5239\pm0.1113$ & $1.4304$ & $0.1107\pm6e^{-6} $ & $0.0016$\\
		\hline

		32 & $2.4639\pm0.0220$ & $2.5034$ & $2.6133\pm0.0245$ & $2.6448$ & $0.0016\pm8e^{-6}$ & $0.0016$\\
		\hline
    
		64 & $4.7786\pm0.0485$ & $4.8106$ & $4.9209\pm0.0238$ & $4.9572$ & $0.0013\pm{6e^{-6}}$ & $0.0013$\\
		\hline

		128 & $9.1063\pm0.0659$ & $9.1122$ & $9.4467\pm0.0282$ & $9.4117$ & $0.0014\pm4e^{-6}$ & $0.0014$\\
		\hline

		256 & $33.9273\pm1.0659$ & $35.3962$ & $20.6083\pm0.4611$ & $20.7819$ & $0.0014\pm5e^{-6}$ & $0.0014$\\
		\hline

		512 & $74.3700\pm3.6795$ & $65.0996$ & $74.5755\pm3.1540$ & $67.4778$ & $0.0014\pm6e^{-6}$ & $0.0014$\\
		\hline

		1024 & $105.1532\pm3.4044$ & $95.7272$ & $98.4837\pm1.1094$ & $96.5968$ & $0.0014\pm5e^{-6}$ & $0.0014$\\
		\hline

		2048 & $-$ & $-$ & $-$ & $-$ & $0.0014\pm4e^{-6}$ & $0.0014$ \\
		\hline

		4096 & $-$ & $-$ & $-$ & $-$ & $0.0014\pm4e^{-6}$ & $0.0014$\\
		\hline

		8192 & & $-$ & $-$ & $-$ & $0.0014\pm5e^{-6}$ & $0.0014$\\
		\hline

        16384 & $-$ & $-$ & $-$ & $-$ & $0.1652\pm2e^{-5}$ & $0.0014$\\
		\hline
        
	\end{tabular}
    \vspace{0.3cm}
	\caption{Performance Comparison of Optimizers for One-Step Ahead Prediction. The table shows the time taken by Adam, DNI-Adam, and Dopamine 2 optimizers for the \textbf{backward pass} during one-step ahead prediction tasks in seconds(wall clock time).} 
	\label{tab:time_epoch_backward}
\end{table}

\begin{table}[h]
	\centering
	\begin{tabular}{c|c|c|c|c|c|c|}
		Look-back window & \multicolumn{2}{c|}{Adam} & \multicolumn{2}{c|}{DNI-Adam} & \multicolumn{2}{c|}{Dopamine 2} \\
        \cline{2-7}
        & Mean ± SEM & Median & Mean ± SEM & Median & Mean ± SEM & Median \\
		\hline 
  
		16 & $2.6805\pm0.2201$ & $2.4720$ & $2.9312\pm0.0160$ & $2.5736$ & $0.6571\pm0.0022 $ & $0.6560$\\
		\hline
  
		32 & $3.9528\pm0.0170$ & $3.9585$ & $4.2859\pm0.0282$ & $4.2731$ & $0.8005\pm0.0057$ & $0.7973$\\
		\hline

		64 & $7.2394\pm0.0192$ & $7.2186$ & $7.7260\pm0.0485$ & $7.6966$ & $1.0639\pm0.0071$ & $1.0485$\\
		\hline

		128 & $14.0686\pm0.0259$ & $14.0307$ & $14.7559\pm0.0811$ & $14.7426$ & $1.7080\pm0.0217$ & $1.7024$\\
		\hline

		256 & $43.7883\pm1.0899$ & $45.7838$ & $30.5137\pm0.4312$ & $30.0077$ & $2.9480\pm0.0356$ & $2.9806$\\
		\hline

		512 & $93.5957\pm3.6968$ & $84.1061$ & $94.0851\pm3.2305$ & $86.8482$ & $4.7308\pm0.0335$ & $4.6767$\\
		\hline

		1024 & $143.3164\pm3.2120$ & $135.2638$ & $139.3597\pm0.7855$ & $138.4504$ & $9.0808\pm0.0890$ & $8.9271$\\
		\hline

		2048 & $-$ & $-$ & $-$ & $-$ & $16.6741\pm0.0175$ & $16.6894$ \\
		\hline

		4096 & $-$ & $-$ & $-$ & $-$ & $31.2805\pm0.0402$ & $31.3420$\\
		\hline

		8192 & & $-$ & $-$ & $-$ & $56.6109\pm0.0814$ & $56.7139$\\
		\hline

        16384 & $-$ & $-$ & $-$ & $-$ & $91.1938\pm0.1003$ & $91.1241$\\
		\hline
        
	\end{tabular}
    \vspace{0.3cm}
	\caption{Performance Comparison of Optimizers for One-Step Ahead Prediction. The table shows the time taken by Adam, DNI-Adam, and Dopamine 2 optimizers for both \textbf{forward \& backward pass} during one-step ahead prediction tasks in seconds.} 
	\label{tab:time_epoch_all}
\end{table}





\end{document}